\documentclass{article} 
\usepackage{collas2025_conference,times}
\usepackage{easyReview}

\usepackage{amsmath,amsfonts,bm}

\def\eqref#1{equation~\ref{#1}}

\def\1{\bm{1}}

\DeclareMathAlphabet{\mathsfit}{\encodingdefault}{\sfdefault}{m}{sl}
\SetMathAlphabet{\mathsfit}{bold}{\encodingdefault}{\sfdefault}{bx}{n}

\usepackage{hyperref}
\hypersetup{
    colorlinks=true,
    linkcolor=red,
    filecolor=magenta,
    urlcolor=blue,
    citecolor=purple,
    pdftitle={Overleaf Example},
    pdfpagemode=FullScreen,
    }

\usepackage{multirow}
\usepackage{graphicx}
\usepackage{caption}
\usepackage{subcaption}
\usepackage{array}
\usepackage{arydshln}
\usepackage{makecell}
\usepackage{booktabs}
\usepackage{algorithm}
\usepackage{algpseudocode}
\usepackage{threeparttable}
\usepackage{wrapfig}

\title{CLA: Latent Alignment for Online Continual Self-Supervised Learning}

\author{Giacomo Cignoni\\
University of Pisa\\
Largo B. Pontecorvo 3, Pisa, Italy\\
\texttt{giacomo.cignoni@phd.unipi.it}\\
\And
Andrea Cossu\\
University of Pisa\\
Largo B. Pontecorvo 3, Pisa, Italy\\
\texttt{andrea.cossu@unipi.it}\\
\And
Alexandra Gomez-Villa\\
Computer Vision Center (CVC)\\
Edifici O, 08193 Bellaterra, Barcelona\\
\texttt{agomezvi@cvc.uab.es}\\
\And
Joost van de Weijer\\
Computer Vision Center (CVC)\\
Edifici O, 08193 Bellaterra, Barcelona\\
\texttt{joost@cvc.uab.es}\\
\And
Antonio Carta\\
University of Pisa\\
Largo B. Pontecorvo 3, Pisa, Italy\\
\texttt{antonio.carta@unipi.it}\\
}

\collasfinalcopy 

\preprintcopy 

\begin{document}

\maketitle

\begin{abstract}
Self-supervised learning (SSL) is able to build latent representations that generalize well to unseen data. However, only a few SSL techniques exist for the online CL setting, where data arrives in small minibatches, the model must comply with a fixed computational budget, and task boundaries are absent. We introduce Continual Latent Alignment (CLA), a novel SSL strategy for Online CL that aligns the representations learned by the current model with past representations to mitigate forgetting. We found that our CLA is able to speed up the convergence of the training process in the online scenario, outperforming state-of-the-art approaches under the same computational budget. Surprisingly, we also discovered that using CLA as a pretraining protocol in the early stages of pretraining leads to a better final performance when compared to a full i.i.d. pretraining. Code available at: \url{https://github.com/giacomo-cgn/cla} .
\end{abstract}

\section{Introduction}
\label{sec:intro}

\begin{figure}[b]
    \centering
    \begin{subfigure}{0.36\textwidth}
      \centering
      \includegraphics[width=1.0\linewidth]{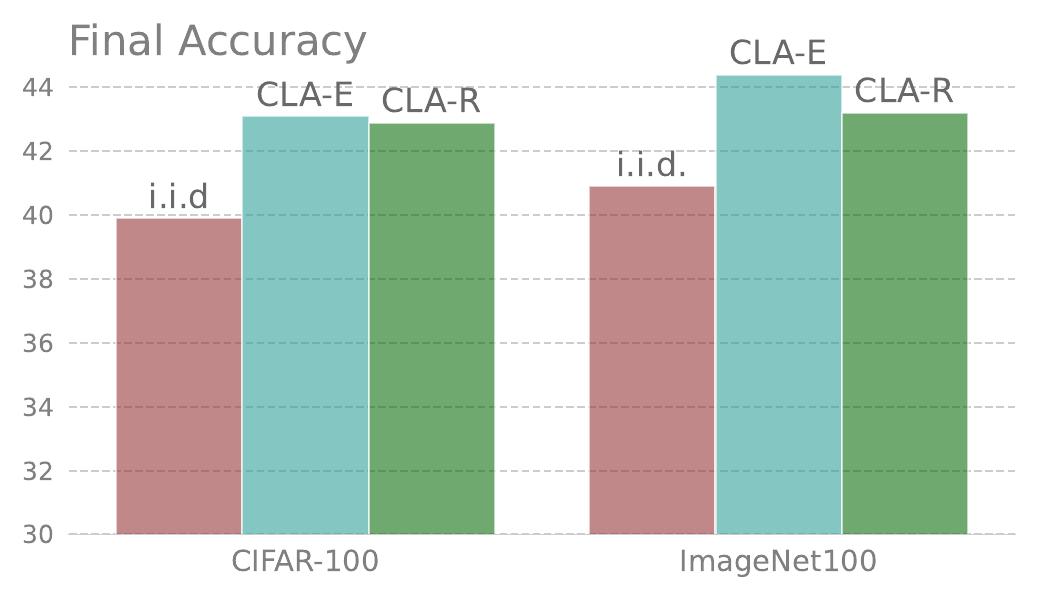}
    \end{subfigure}
    \begin{subfigure}{0.63\textwidth}
      \centering
      \includegraphics[width=1.0\linewidth]{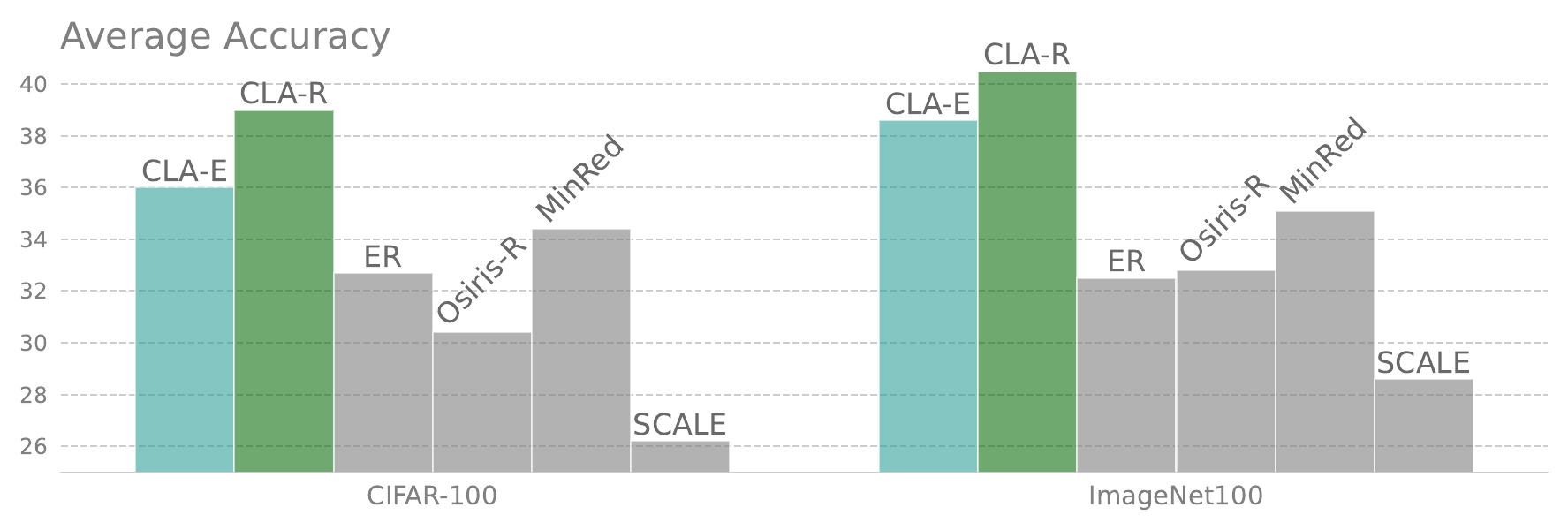}
    \end{subfigure}
    
    \caption{Notable results: when trained with the same computational budget (CBP), CLA outperforms i.i.d. on the \textsc{Final Accuracy}. CLA also outperforms state-of-the-art CL-SSL methods on the \textsc{Average Accuracy} (average of accuracies evaluated after each experience, indicates performance across the entire training stream). }
    \label{fig:barplots}
\end{figure}

Self-Supervised Learning (SSL) has proven to be an effective paradigm for learning high-quality representations from unlabeled data \citep{chen2020simsiam, chen2020simclr, grill2020byol, zbontar2021barlow, caron2021emerging}.
However, the majority of SSL methods assume that all training data is jointly available, which is not realistic for many applications where data from drifting distributions arrives over time.

Continual Learning (CL) relaxes this assumption by considering algorithms that can learn from non-stationary data streams \citep{lesort2019continual, kirkpatrick2017ewc, goodfellow2015empirical, gomez2024exemplar}. 
Even though SSL is generally considered more resilient to forgetting of previous knowledge \citep{cossu2022pretraining}, SSL approaches for CL work in an ``offline'' setting, where the data stream is partitioned into a set of well-defined tasks with known boundaries \citep{fini2022cassle, gomezvilla2022pfr, cha2023sycon}. Furthermore, these methods do not comply with a limited computational budget, and they are trained on each task for as long as needed.

We focus on the challenging ``online'' CL scenario \citep{mai2021onlinesurv, soutifcormerais2023comprehensive,lopezpaz2022gem}, where information about task boundaries is not available and the model is trained in one pass on the incoming data stream, with only few data samples available at a time. The one-pass training protocol and the small minibatch size require fast adaptation and quick convergence, since the model can only afford a few training iterations on a given set of examples before new ones arrive. 
This is of particular interest for SSL, as their performance usually depends on a large computational budget.
We will refer to the online CL scenario with SSL methods as \emph{OCSSL}. We emphasize the importance of OCSSL, as real-world environments often lack labeled data and demand rapid adaptation and anytime evaluation \citep{soutifcormerais2023comprehensive, caccia2021new, rebuffi2017icarl, ridge2015robot}. 
Studying OCSSL solutions can also offer a different perspective on the behavior of CL strategies, by highlighting the importance of fast adaptation, rather than focusing almost exclusively on the mitigation of forgetting.
Given these considerations, we identify the aim of OCSSL as training a feature extractor capable of retaining meaningful representations for all data in the stream. An important desideratum is fast convergence, as the length of the training stream is typically unknown, meaning the model should be useful for inference at any point during the training process. We measure this capability through \textsc{Average Accuracy}.

We propose a novel strategy for OCSSL named Continual Latent Alignment (CLA), that aligns the representations learned by the current model with past representations. CLA achieves state-of-the-art performance over the entire training trajectory, beating other methods in \textsc{Average Accuracy} (Fig. \ref{fig:barplots}), without relying on external information about task boundaries. To ensure fair comparisons across different OCSSL methods, we introduce a new metric called Cumulative Backward Passes (CBP) and we only evaluate models sharing the same CBP. \emph{CLA is able to surpass the performance of i.i.d. training for relatively small values of CBP} (Figure \ref{fig:barplots}), which makes it a very effective strategy for OCSSL. We hypothesize that this is due to CLA exhibiting fast convergence abilities, especially during the early stages of training. To validate this hypothesis we show that CLA can be used as a more effective pretraining method than i.i.d. when applied in the first pretraining stages: continuing i.i.d. pretraining from a CLA-based initialization leads to better results than with a full i.i.d. training. 
We summarize our main contributions:
\begin{itemize}
    \item We propose \textit{Continual Latent Alignment} (CLA), a strategy for OCSSL based on learned alignment of latent representations.
    \item We define a metric (CBP) that allows a fair comparison of OCSSL methods with the same computational budget.
    \item We conduct thorough experiments of novel and existing CL strategies that are applicable in OCSSL and verified that CLA reaches state-of-the-art performance.
    \item We show that CLA exhibits fast adaptation capabilities and that CLA itself can be used as a pretraining protocol during the early stages of pretraining.
\end{itemize}

\section{Related Works}
\paragraph{\textsc{Self-Supervised Learning}}
SSL methods train a feature extractor network, $\theta: \mathcal{X} \rightarrow \mathcal{F}$ to map input $x \in X$ to a latent representation $z$ in the feature space $\mathcal{F}$. Training happens on pretext tasks, which are objectives specifically designed to learn useful representations without labels \citep{gui2023sslsurvey, ericsson2022sslintro}, and evaluation is usually performed via linear probing on classification tasks \citep{alain2018probing}.
We are interested in SSL methods using \textit{instance discrimination} pretext tasks \citep{gui2023sslsurvey}: two different views, $x_1$ and $x_2$ are extracted from the same sample and the pretext task consists in enforcing the two encoded views to be close in the feature space.
Contrastive methods like SimCLR \citep{chen2020simclr} and MoCo \citep{he2020moco, chen2020mocov2} are instance discrimination methods that apply data augmentation to create positive pairs of a given example, while treating other batch samples as negatives.
Other methods, instead, avoid the explicit use of negative examples \citep{zbontar2021barlow, caron2021swav, bardes2022vicreg, caron2019deepcluster}, notable examples being SimSiam \citep{chen2020simsiam}, BYOL \citep{grill2020byol} and DINO \citep{caron2021dino}, which all use an additional predictor head and enforce high similarity of features coming from the same example. 

\paragraph{\textsc{Continual Self-Supervised Learning}}
Mitigating the forgetting of previous knowledge is one of the major objectives of CL approaches \citep{chen2016lifelong, kirkpatrick2017ewc, MCCLOSKEY1989109, lesort2019continual}.
Continual Self-Supervised Learning (CSSL) introduces the continual element in SSL by considering a non-stationary data stream $\mathcal{D} = (\mathcal{D}_1, ..., \mathcal{D}_t, ..., \mathcal{D}_T)$.
In the popular class-incremental setting \citep{vandeven2019scenarios}, the learner faces a sequence of $T$ tasks, each with its own data distribution $\mathcal{D}_t$. Each $\mathcal{D}_t$ contains samples belonging to a set of classes $\mathcal{Y}_t$ such that $\mathcal{Y}_t \cap \mathcal{Y}_s = \emptyset$ for each other task $s \neq t$. Note that in the CSSL setting, the class label is unknown to the model and is only used to split the dataset. The CSSL model is trained sequentially on the data stream.
In addition to CSSL, various papers have explored label-scarce scenarios in a CL setting, i.e. continual semi-supervised learning \citep{brahma2021hypernetworkscontinualsemisupervisedlearning, smith2021memoryefficientsemisupervisedcontinuallearning}. In these scenarios, the inclusion of SSL techniques has proven beneficial in preventing forgetting \citep{pham2021dualnet}, especially with an online data stream \cite{gallardo2021selfsupervisedtrainingenhancesonline}.

Most CSSL strategies are applied to \textit{instance discrimination} SSL methods and assume an offline CL scenario, i.e. the model can reach convergence with multi-epoch training on each experience, with well defined task boundaries.
A popular CSSL solution is employing the frozen encoder at the end of the previous task as teacher for feature distillation. This is the case for both CaSSLe \citep{fini2022cassle} and PFR \citep{gomezvilla2022pfr}, which we discuss in more detail in Section \ref{sec:motivations}, as we follow a similar paradigm in CLA.
SyCON \citep{cha2023sycon} and POCON \citep{gomezvilla2023pocon} use similar distillation approaches, with a contrastive regularization loss and a pair of fast and slow networks, respectively.
Osiris-D and Osiris-R \citep{zhang2024osiris} leverage the NT-Xent loss \citep{sohn2016ntxent} and use a cross-task symmetric loss that contrasts current and replay memory samples. Osiris-D combines this with distillation from past task frozen model, while Osiris-R computes its SSL loss on replayed samples.
 LUMP \citep{madaan2022lump}, instead, performs training on samples generated by interpolating each current task sample with a memory exemplar.
MinRed \citep{purushwalkam2022minred} only uses memory exemplars in the SSL loss, while keeping in the memory a subset of maximally decorrelated samples, with sample correlation calculated from their network feature representation.
\cite{cignoni2025cmp} explore a multipatch approach for replay-free CSSL.
Overall, we find that existing CSSL methods are not crafted to address issues found in OCSSL, thus, most of them cannot be directly applied in this scenario: they often require task boundaries and they do not take into account a given computational budget.

SCALE \citep{yu2023scale} is the only existing OCSSL strategy. Instead of enhancing an underlying SSL method, it includes its own SSL loss consisting of an InfoNCE-like loss \citep{oord2019infonce} and a forgetting-prevention component based on the KL distance between the current and last training step representations pairwise similarity. It also employs a Part and Select Algorithm \citep{salomon2013psa} updated buffer. We compare our CLA with SCALE, while also focusing on the two desiderata of fast adaptation under a fixed computational budget.

\section{OCSSL Scenario}
\label{sec:ocssl}

\begin{figure}[ht]
    \centering
    \includegraphics[width=\linewidth]{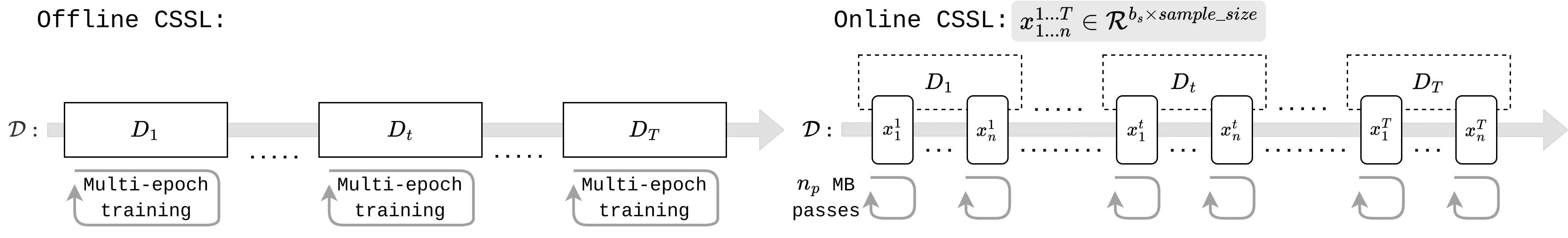}
    \caption{Compared to offline CSSL, in OCSSL data is presented as a one-pass continual stream, with no explicit task boundaries, data incoming in the form of small minibatches, and without the possibility of multi-epoch training (although executing $n_p$ consecutive passes on the same minibatch is permitted).}
    \label{fig:cl-streams}
\end{figure}

The OCSSL scenario shares many similarities with the online (supervised) CL scenario \citep{soutifcormerais2023comprehensive, mai2021onlinesurv}. However, examples in an OCSSL stream do not come with a pre-defined target label. Importantly, the data is presented to the model in a single one-pass stream, one (usually small) minibatch at a time, without the ability of revisiting past minibatches when a new one has arrived. Therefore, multi-epoch training is prevented in OCSSL, although one can iterate multiple ($n_p$) times on the same minibatch before passing to the next one.
Thus, we can identify two levels of abstractions for the subdivision of the data stream $\mathcal{D}$: the first is the subdivision of the entire stream $\mathcal{D}$ in class-incremental distributions $\mathcal{D}_t$, this subdivision is unknown to the model as task boundaries are not explicit; the second is the subdivision in each minibatch $x_i^t \in \mathcal{D}_t$, each of limited size $b_s$.
Fig. \ref{fig:cl-streams} illustrates the differences between offline CSSL and OCSSL.
The difficulties of the OCSSL scenario, compared to conventional offline CSSL can be identified as: (1) absence of explicit task boundaries; (2) small streaming minibatches (with fixed size $b_s$), as this is particularly relevant for SSL methods that benefit from large minibatch sizes \citep{chen2020simclr, zbontar2021barlow}; (3) the one-pass stream, both because it is difficult to ensure convergence during training and because the learner cannot revisit previous minibatches.

\subsection{Motivations}
\label{sec:motivations}
Knowledge distillation of latent features is used in CSSL to prevent forgetting. In this section, we first present two popular approaches from the literature, PFR \citep{gomezvilla2022pfr} and CaSSLe \citep{fini2022cassle}, and we generalize their formulation, which will be used to derive our proposed method in the next section.
Both CaSSLe and PFR  enhance an underlying SSL method for a CL scenario by introducing a loss regularization term $\mathcal{L}_\textit{reg}$, obtained by performing distillation of the features $z: \mathcal{Z}$ of the current network $\theta : \mathcal{X} \rightarrow \mathcal{Z}$, passed through an additional projector, using as targets the features $z^{t-1} : \mathcal{Z}$ of a past network $\theta^{t-1}$ frozen at the last task boundary.
We call this process \emph{alignment} and we name the additional projector $a_\phi$.
In more detail, the alignment operation consists in passing $z$ through $a_\phi : \mathcal{Z} \rightarrow \mathcal{Z}$, and then enforcing similarity to $\hat{z}$ through an alignment loss $\mathcal{L}_\textit{alg}$. 
The inclusion of $a_\phi$ is to map the features learned on the current experience back to the past feature space; this is done to maintain plasticity for the current feature space.
We can define the loss formulation that encompasses both CaSSLe and PFR as:
\begin{equation}
\label{eq:reg}
    \mathcal{L} = \mathcal{L}_\textit{SSL}(z_1, z_2) + \omega \mathcal{L}_\textit{reg} \ , \ \ 
 \ \mathcal{L}_\textit{reg} = \frac{\mathcal{L}_\textit{alg}(a_\phi (z_1), z_1^{t-1})}{2} +
    \frac{\mathcal{L}_\textit{alg}(a_\phi(z_2), z_2^{t-1})}{2} \ ,
\end{equation}
where $z_1 = \theta(x_1)·,\ z_2=\theta(x_2),\ z_1^{t-1} = \theta^{t-1}(x_1),\ z_2^{t-1}=\theta^{t-1}(x_2)$.
It is important to note that CaSSLe and PFR differ, as the first uses $\mathcal{L}_\textit{alg}=\mathcal{L}_\textit{SSL}$ and a fixed regularization strength $\omega=1$, while the latter uses negative cosine similarity for alignment, $\mathcal{L}_\textit{alg}=-S_C$. The alignment is performed on the features extracted before the projector network. CaSSLe, instead, uses the features after the projector. 

When applied to an OCSSL scenario, both strategies present two main limitations:
\begin{enumerate}
    \item They require explicit task boundaries, as they use the output features of the frozen encoder at task boundaries as targets for the alignment.
    \item They do not incorporate replay strategies. In the OCSSL scenarios, the small minibatch size is a strong constraining factor for the performance, and rehearsal techniques are crucial to increase the minibatch size while also including information from the past to mitigate forgetting. 
\end{enumerate}

\section{CLA: Continual Latent Alignment}
\label{sec:cla}

\begin{figure}[tb]
    \centering
    \begin{subfigure}{0.25\textwidth}
      \centering
      \includegraphics[width=1.0\linewidth]{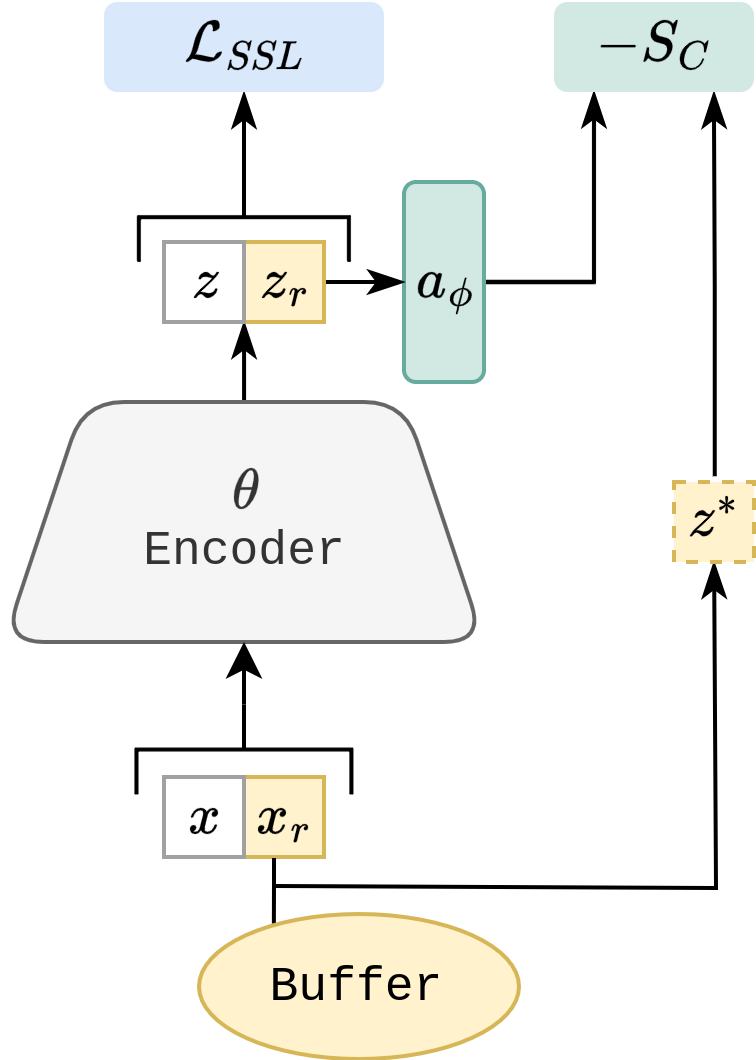}
      \caption{CLA-R}
    \end{subfigure}
    \hspace{1cm}
    \begin{subfigure}{0.35\textwidth}
      \centering
      \includegraphics[width=1.0\linewidth]{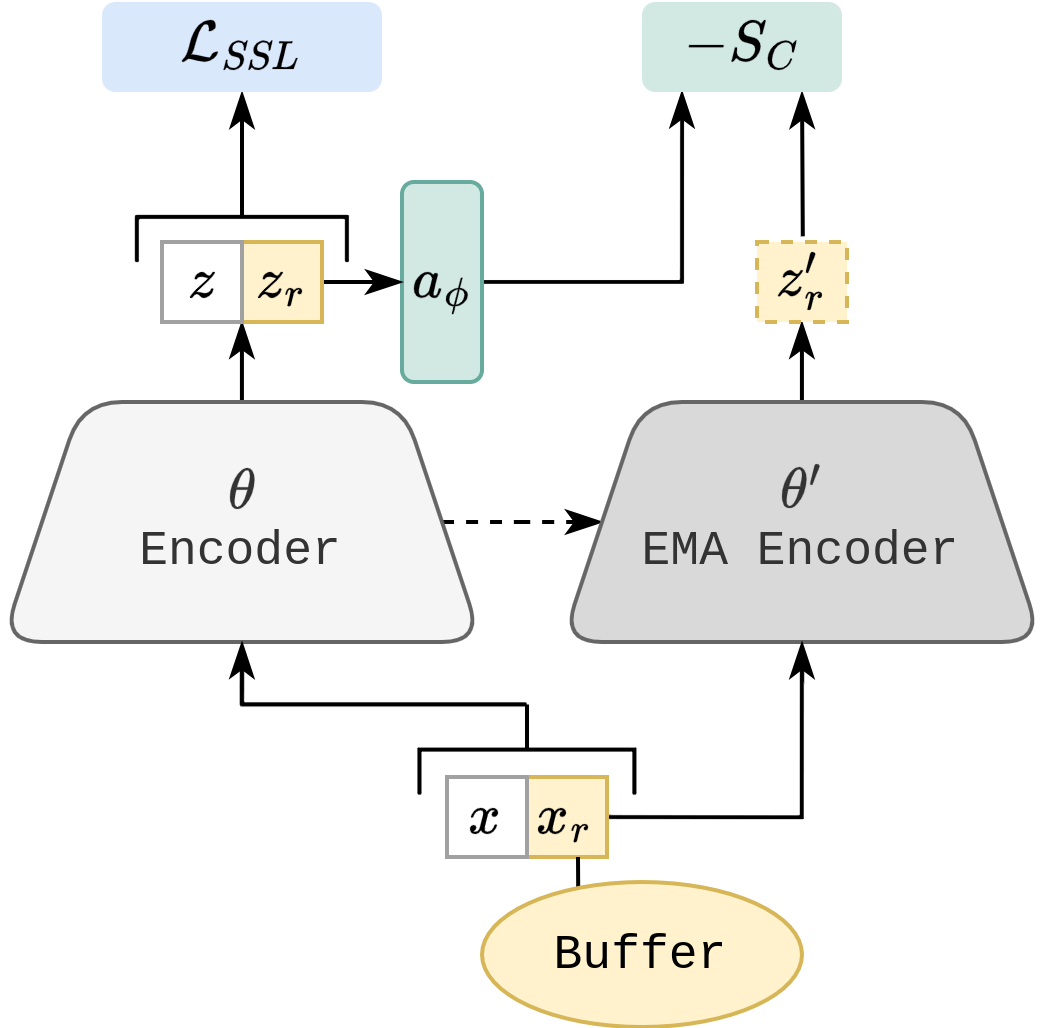}
      \caption{CLA-E}
    \end{subfigure}
    
    \caption{Overview of the novel strategies based on alignment to past representations for OCSSL.}
    \label{fig:novel-strategies}
\end{figure}

\textbf{\textit{Continual Latent Alignment} (CLA)} is a novel strategy for OCSSL. It addresses the first aforementioned limitation by performing alignment based on an Exponential Moving Average (EMA) network $\theta'$. The EMA network does not need task boundary information to be updated. 
We call this strategy \textbf{CLA-b}, the \emph{base} variant of CLA without rehearsal.
The EMA update for the network is defined by $\theta' \ \leftarrow \ \tau \theta' + (1-\tau)\theta $ and regulated by the scalar parameter $\tau$. The network is updated after each minibatch. The benefits of EMA networks for a CL setting have also been pointed out in \citep{soutifcormerais2023improving}. 
The idea behind CLA-b is that enforcing current features to be aligned to an EMA network helps to maintain past representations and stabilizes the training process.
In fact, \cite{michel2024rethinkingmomentumknowledgedistillation} proved the positive effects on distillation (which is conceptually similar to alignment) from an EMA network in an Online CL setting, with advantages in reducing feature drift, improving feature discrimination and backward transfer. These effects are especially important in the early phases of training and help speed up convergence compared to models that do not employ such measures and are thus fully subject to the instabilities of the OCSSL stream.

The CLA-b loss reads:
\begin{equation}
        \mathcal{L}_\textit{reg\_CLA-b} = \frac{-S_C(a_\phi (z_1), z_1')}{2} +
    \frac{-S_C(a_\phi(z_2), z_2')}{2} \ ,
    \end{equation}
    where $-S_C$, the negative cosine similarity, is the alignment loss $\mathcal{L}_\textit{alg}$ that considers $z_{1}=\theta(x_{1})$, $ \ z_{2}=\theta(x_{2})$, $ \ z_{1}'=\theta'(x_{1})$, $ \ z_{2}'=\theta'(x_{2})$. The functions $\theta$ and $\theta'$ include both the encoder and the projector network.

Starting from CLA-b, we now introduce replay to account for the second limitation mentioned before. Replay exemplars $x_r$ are concatenated to streaming batch samples, resulting in two views after the augmentations: $x_2 \cup x_{r2}$ and $x_1 \cup x_{r1}$.
In this case, we define $z_1 \cup z_{r1}$, $z_2 \cup z_{r2}$ as the encoder output features for both sample views, which are composed by the concatenation of the streaming sample features ($z_1, z_2$) and the replay samples features ($z_{r1}, z_{r2}$).
In order to grant more plasticity for the novel tasks, only replay features $z_{r1}$ and $z_{r2}$ are used in the alignment process, while the SSL loss uses both stream and replay samples features.

We show two variants of CLA that only differs in the origin of the target features:
\begin{itemize}
    \item \textbf{CLA-E}, similarly to CLA-b uses the output of an EMA-updated network $\theta'$ as the target features for alignment.
    \item \textbf{CLA-R}, stores past network output features in the memory buffer in pair with the original samples. Replay samples $x_r$ are extracted from the buffer together with the corresponding past features $z^*$, which is used as the alignment target for $z_{r1}, z_{r2}$. 
\end{itemize}

We finally define the CLA losses as:
\begin{equation}
\begin{split}
        \mathcal{L}_\textit{CLA-E/R} &= \mathcal{L}_\textit{SSL}(z_1 \cup z_{r1}, z_2 \cup z_{r2}) + \omega \mathcal{L}_\textit{reg\_CLA-E/R} \ ,\\
        \mathcal{L}_\textit{reg\_CLA-E} &= \frac{-S_C(a_\phi (z_{r1}), \hat{z}_1)}{2} + \frac{-S_C(a_\phi(z_{r2}), \hat{z}_2)}{2} \ , \\
        \mathcal{L}_\textit{reg\_CLA-R} &= \frac{-S_C(a_\phi (z_{r1}), z^*)}{2} + \frac{-S_C(a_\phi(z_{r2}), z^*)}{2} \ .
\end{split}
\end{equation}
where $ \hat{z}_1 = \theta'(x_{r1})$, $\hat{z}_2 = \theta'(x_{r2})$, $z^* \sim \mathcal{M}$. We report the pseudocode for all CLA variants in Appendix \ref{sec:pseudocode}.

\paragraph{\textsc{FIFO Buffer}}
We chose to use a FIFO buffer in CLA-E and CLA-R: exemplars in the buffer are updated with a First-In-First-Out strategy, while sampling for rehearsal remains random. For CLA-R, at the end of each training step, the past target feature $z^*$ in the buffer is updated with $z^* = 0.5 \cdot z^* + 0.5 \cdot \frac{z_{r1}+z_{r2}}{2}$.
The buffer has an important impact on the convergence of the method. A reservoir buffer \citep{vitter1985random} provides an unbiased sample of the stream, but results in an uneven number of iterations for each sample and higher overfitting on the buffer samples. Instead, with a FIFO buffer the expected number of training iterations for each sample in the stream is equal, thus inducing better convergence in exchange for a stronger bias towards recent experiences, which could induce more forgetting.
Coherently with our hypothesis of convergence being fundamental in OCSSL, we argue that, in this scenario, convergence benefits, can overcome the forgetting disadvantages, especially when there are other components of the strategies (i.e. the alignment loss) that prevent forgetting.

\section{CBP: a Fair Metric for OCSSL}
\label{sec:cbp}
The concept of comparing Online CL strategies with an equal computational budget is already present in the literature \citep{prabhu2023oclstorageconstraint, prabhu2023computbudgetedcl}, also by including the training time \citep{wang2024forgettingignorancemyopiarevisiting}.

We adopt the same approach for OCSSL: in a scenario where performance is highly dependent on fast convergence, we decided to use the number of backward passes as a proxy for the computational budget.
Thus, we define the \textit{\textbf{Cumulative Backward Passes}} (\textbf{CBP}) metric for OCSSL:
\begin{equation}
        \text{CBP} = n_v \times n_\text{steps} \times b \ ,
\end{equation}
where $n_v$ is the number of views needed by the SSL method, $b$ is the minibatch size, and $n_\text{steps}$ is the total number of training steps. We compute $n_\text{steps}$ as:
\begin{equation}
    \begin{split}
        n_\text{steps} &= n_p \times \frac{N}{b_s} ,
    \end{split}
\end{equation}
where $b_s$ is the streaming minibatch size, $N$ the total number of samples in the training dataset and $n_p$ the number of sequential minibatch training passes executed for each incoming streaming minibatch.\\
\begin{wrapfigure}{r}{0.59\textwidth}
    \centering
    \includegraphics[width=1.0\linewidth]{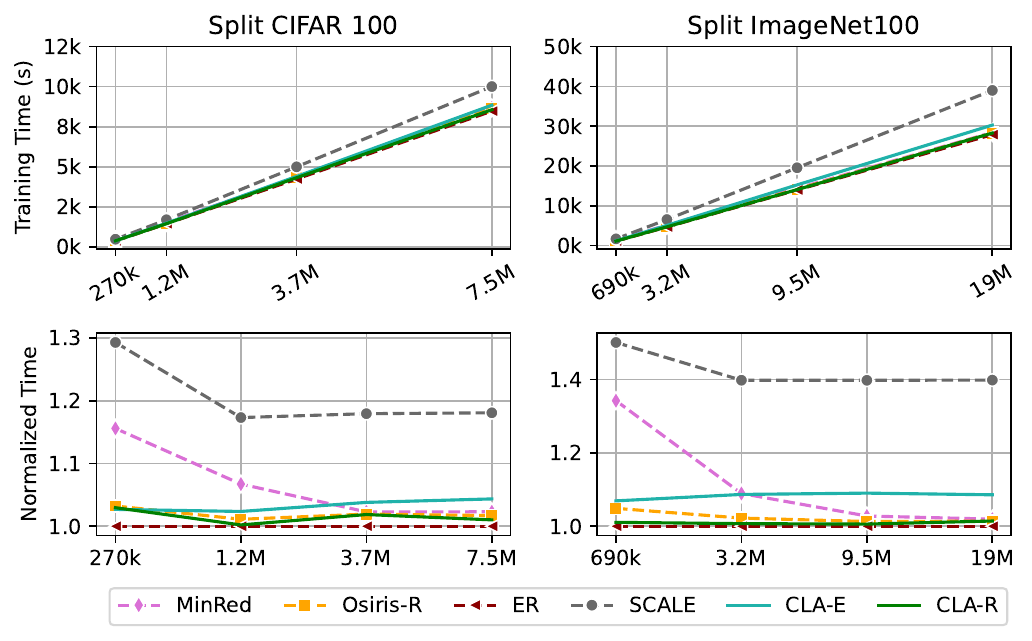}
    \caption{Top row: absolute training times in seconds with varying CBP. Bottom row: normalized training times relative to the best time.}
    \label{fig:cbp-times}
\end{wrapfigure}
CBP is intended as a measure of the ``abstract'' computational cost: it does not exactly measure the computation time, as it abstains from considering hardware-related bottlenecks and implementation differences. This is intentional as it is a measure that does not depend on the underlying hardware infrastructure, it is easy to compute and should not lead to reproducibility issues.

With CBP, there are two ways to increase the computational budget: one is to increase the minibatch size $b$ with replay samples, the other is to increase $n_p$, the number of training passes per minibatch. Usually, OCSSL strategies with larger minibatches (but same number of $n_\text{steps}$) obtain better results \citep{yu2023scale}. Therefore, it is crucial to fix CBP when comparing different strategies. The composition of the minibatch is another important factor as it determines the minibatch size parameter $b$: it can be composed by only stream samples ($b=b_s$), only by replay samples ($b=b_r$) or by a combination of both ($b=b_r + b_s$). We report the composition of minibatches for each method in Tables \ref{tab:results-high-cbp} and \ref{tab:results-low-cbp}.

Figure \ref{fig:cbp-times} reports training times of methods for varying CBP. In the absolute time plots, it is easy to see the soundness of our metric, as execution times scale linearly with the CBP and all methods employ similar times. 
The limitations of our CBP metric include its focus on computational burdens associated with backward passes, which may not fully account for other sources of overhead. For instance, as shown in the relative time plots in Figure \ref{fig:cbp-times}, SCALE incurs small additional computational cost due to its complex loss function and the need to copy the encoder at each training step. Similarly, CLA-E requires an extra EMA forward pass, leading to a very slight increase in computational time compared to simpler baselines like ER. However, these differences remain significantly modest for the methods analyzed.

\paragraph{\textsc{i.i.d. Training}} Comparing novel and existing methods to the offline i.i.d. upper bound involves removing both the \emph{continual} and \emph{online} components from OCSSL, while maintaining the same computational budget in terms of CBP. For the i.i.d. baseline, we train the plain SSL method on all tasks simultaneously (using an i.i.d. distribution of the entire dataset), with multiple epoch training and setting the minibatch size to the same $b$ used by the OCSSL strategies.
The number of epochs that guarantees equivalent CBP for both i.i.d. and OCSSL strategies is:
\begin{equation}
\label{eq:epochs_iid}
    n_\text{epochs\_iid} = \Bigl\lceil n_p \times \frac{b}{b_s} \Bigr\rceil \ .
\end{equation}

\section{Experiments}
\label{sec:experiments}

\begin{table}[b]
    \centering
    \caption{\textbf{Results for High CBP} on Split-CIFAR100 and Split-Imagenet100, with CBP = 3.7M for CIFAR-100 and CBP = 9.5M for ImageNet100, using $b_s=10$, $b=138$, $n_p=3$. Best in \textbf{bold}, second best \underline{underlined}.}
    \label{tab:results-high-cbp}
    \begin{tabular}{ccccccc}
    \toprule
          & & & \multicolumn{2}{c}{CIFAR100 (20 exps)} & 
         \multicolumn{2}{c}{ImageNet100 (20 exps)} \\
         \multicolumn{2}{c}{\textsc{Method}} & $b$ \textsc{composition} & \textsc{Final Acc.} & \textsc{Avg. Acc.} & \textsc{Final Acc.} & \textsc{Avg. Acc.} \\
         \cmidrule(rl){1-2}\cmidrule(rl){3-3}\cmidrule(rl){4-5} \cmidrule(rl){6-7}
         \multirow{7}{*}{\rotatebox[origin=c]{90}{\textsc{SimSiam}}} 
         & \textit{i.i.d.} & - & $39.9 \pm {\scriptstyle 0.4}$ & - & $40.9 \pm {\scriptstyle 0.8}$ & - \\
         & \textit{CaSSLe-R} & $b=b_s+b_r$ & $40.8 \pm {\scriptstyle 0.2}$ & $35.2 \pm {\scriptstyle 0.1}$ & $38.8 \pm {\scriptstyle 0.3}$ & $36.1 \pm {\scriptstyle 0.2}$ \\
         & ER w/ reservoir & $b=b_s+b_r$ & $41.5 \pm {\scriptstyle 0.1}$ & $32.7 \pm {\scriptstyle 0.1}$ & $40.8 \pm {\scriptstyle 0.9}$ & $32.5 \pm {\scriptstyle 0.3}$  \\
         & ER w/ FIFO & $b=b_s+b_r$ & $40.5 \pm {\scriptstyle 0.7}$ & $32.1 \pm {\scriptstyle 0.2}$ & $38.0 \pm {\scriptstyle 1.2}$ & $31.3 \pm {\scriptstyle 0.4}$  \\
         & MinRed & $b=b_r$ & $41.8 \pm {\scriptstyle 0.0}$ & $34.4 \pm {\scriptstyle 0.1}$ & $\underline{44.2 \pm {\scriptstyle 0.2}}$ & $35.1 \pm {\scriptstyle 0.4}$  \\
         & \textbf{CLA-R} & $b=b_s+b_r$ & $\underline{42.9 \pm {\scriptstyle 0.3}}$ & $\mathbf{39.0 \pm {\scriptstyle 0.2}}$ & $43.2 \pm {\scriptstyle 0.4}$ & $\mathbf{40.5 \pm {\scriptstyle 0.2}}$  \\
         & \textbf{CLA-E} & $b=b_s+b_r$ & $\mathbf{43.1 \pm {\scriptstyle 0.3}}$ & $\underline{36.0 \pm {\scriptstyle 0.2}}$ & $\mathbf{44.4 \pm {\scriptstyle 0.9}}$ & $\underline{38.6 \pm {\scriptstyle 0.2}}$  \\
         \midrule
          \multicolumn{2}{c}{SCALE} & $b=b_s+b_r$ & $29.4 \pm {\scriptstyle 0.7}$ & $26.2 \pm {\scriptstyle 0.2}$ & $33.3 \pm {\scriptstyle 0.6}$ & $28.6 \pm {\scriptstyle0.5}$ \\
         \multicolumn{2}{c}{Osiris-R} & $b=b_s+b_r$ & $35.4 \pm {\scriptstyle 0.2}$ & $30.4 \pm {\scriptstyle 0.2}$ & $38.7 \pm {\scriptstyle 0.2}$ & $32.8 \pm {\scriptstyle 0.2}$  \\
         \multicolumn{2}{c}{\textit{random encoder}} & - & $15.4 \pm {\scriptstyle 0.4}$ & - & $14.7 \pm {\scriptstyle 0.7}$ & - \\
         \bottomrule
    \end{tabular}
\end{table}

Experiments were conducted on Split CIFAR-100 \citep{krizhevsky2009cifar} and Split ImageNet100 \citep{deng2009imagenet} class-Incremental benchmarks, with 20 experiences each. We used a streaming minibatch size ($b_s$) of 10 and 3 minibatch passes ($n_p=3$) for each incoming stream minibatch.
The backbone network used for all experiments is ResNet-18 \citep{he2015resnet}. As commonly done in the literature, For CIFAR100 we substituted the first 7x7 convolutional layer with a 3x3 convolutional layer and removed the first MaxPool.
We chose SimSiam as the base SSL method, as it is a simple method, allowing us to compare the efficacy and validate the soundness of the various strategies without the possible interference caused by a more complex $\mathcal{L}_\textit{SSL}$.
We repeated all experiments and analysisis using SimCLR instead of SimSiam in Appendix \ref{sec:simclr-experiments}.
For replay based strategies, we also set the size of the memory buffer to 2000.
In all experiments we used SGD with momentum = 0.9 and weight decay = 1e-4; learning rate was selected with a grid search for each method, using 10\% of the training dataset as validation.
 We also executed a grid search for $\omega$ of CLA, while all the other hyperparameters of the other strategies were obtained from the original implementations.
Hyperparameters are reported and analyzed in Appendix \ref{sec:hyperparams}.
The EMA network update parameter $\tau$ of CLA-b and CLA-E was instead set to 0.999.
Probing is executed with a linear probe trained with a minibatch size of 256 and initial learning rate = 0.05, which decreases by a factor of 3 whenever the validation accuracy stops improving. Training of the probe stops when a minimum learning rate or 100 epochs are reached.

In the same OCSSL scenario characterized by $b_s=10$, we conduct experiments with two different CBP budgets: 
\begin{itemize}
    \item  \textbf{Low CBP budget}, Table~\ref{tab:results-low-cbp}, we use CBP = 270k for CIFAR-100 and CBP = 690k for ImageNet100. This translates in having $b=10$ and $n_p=3$ for strategies that do not extend the minibatch size, while having $b=30$ and $n_p=1$, for strategies that use additional replay samples. This allows us to compare both categories of strategies, given the same computational budget.
     \item \textbf{High CBP budget}, Table~\ref{tab:results-high-cbp},  we use CBP = 3.7M for CIFAR-100 and CBP = 9.5M for ImageNet100. In this setting, we only employ strategies that use additional replay samples during training, resulting in $b=138$ and $n_p=3$.
     \end{itemize}
In both settings, we consider the i.i.d. baseline as trained with multiple epochs in an offline fashion, but with the same CBP as the other methods, as explained in Section \ref{sec:cbp}.
We report the results for both final task-agnostic probing accuracy (\textsc{Final Acc.}) and average accuracy of the probing calculated at the end of each experience (\textsc{Avg. Acc.}). This second metric is fundamental: being in an online CL context, it is critical that the model maintains high performance across the entire data stream. For all experiments, we conducted three runs with different seeds and reported mean and standard deviation. 

\begin{table}[bt]
    \centering
    \begin{threeparttable}
    \caption{\textbf{Results for Low CBP} on Split-CIFAR100 and Split-Imagenet100, with CBP = 270k for CIFAR-100 and CBP = 690k for ImageNet100, using $b_s=10$, $b=30$ and $n_p=1$ for \textit{extending MB strategies}, while $b=10$ and $n_p=3$ for \textit{limited MB strategies}. Best in \textbf{bold}, second best \underline{underlined}.}
    \label{tab:results-low-cbp}
    \begin{tabular}{cccccccc}
    \toprule
        & & & \multicolumn{2}{c}{CIFAR100 (20 exps)} & \multicolumn{2}{c}{ImageNet100 (20 exps)} & \\
          \multicolumn{2}{c}{\textsc{Method}} & $b$ \textsc{composition} & \textsc{Final Acc.} & \textsc{Avg. Acc.} & \textsc{Final Acc.} & \textsc{Avg. Acc.}  & \\
         \cmidrule(rl){1-2}\cmidrule(rl){3-3}\cmidrule(rl){4-5} \cmidrule(rl){6-7}
          \multirow{12}{*}{\rotatebox[origin=c]{90}{\textsc{SimSiam}}}  
         & \textit{i.i.d.} & - & $22.1 \pm {\scriptstyle 0.7} $& - & $22.0 \pm {\scriptstyle 0.4}$ & - &
         \multirow{5}{*}{\rotatebox[origin=c]{90}{\footnotesize\textsc{$b=10$ $n_p=3$}}} \\
         & \textit{CaSSLe} & $b=b_s$ & $24.2 \pm {\scriptstyle 0.6}$ & $21.9 \pm {\scriptstyle 0.1}$ & $27.2 \pm {\scriptstyle 0.6}$ & $23.4 \pm {\scriptstyle 0.1}$ & \\
         & finetuning & $b=b_s$ & $21.2 \pm {\scriptstyle 0.2}$ & $20.0 \pm {\scriptstyle 0.1}$ & $21.4 \pm {\scriptstyle 1.}0$ & $20.0 \pm {\scriptstyle 0.1}$ & \\
         & LUMP & $b=b_s=b_r$\tnote{1} & $21.7  \pm {\scriptstyle 0.6}$ & $20.2  \pm {\scriptstyle 0.3}$ & $23.2  \pm {\scriptstyle 0.4}$ & $20.4  \pm {\scriptstyle 0.3}$  & \\
         & \textbf{CLA-b} & $b=b_s$ & $23.9 \pm {\scriptstyle 0.1}$ & $21.7 \pm {\scriptstyle 0.0}$ & $24.6 \pm {\scriptstyle 0.4}$ & $21.1 \pm {\scriptstyle 0.3}$ & \\
         \cmidrule{2-7}
         & \textit{i.i.d.} & - & $22.9 \pm {\scriptstyle 0.2}$ & - & $24.8 \pm {\scriptstyle 0.5}$ & - & 
         \multirow{9}{*}{\rotatebox[origin=c]{90}{\footnotesize\textsc{$b=30$, $n_p=1$}}}\\
         & \textit{CaSSLe-R} & $b=b_s+b_r$ & $26.6 \pm {\scriptstyle 0.5}$ & $23.1 \pm {\scriptstyle 0.1}$ & $25.2  \pm {\scriptstyle 0.4}$ & $21.8 \pm {\scriptstyle 0.3}$ \\
         & ER w/ reservoir & $b=b_s+b_r$ & $21.3 \pm {\scriptstyle 0.2}$ & $20.1 \pm {\scriptstyle 0.1}$ & $25.4 \pm {\scriptstyle 0.3}$ & $22.2 \pm {\scriptstyle 0.4}$ & \\
         & ER w/ FIFO & $b=b_s+b_r$ & $21.4 \pm {\scriptstyle 0.2}$ & $19.9 \pm {\scriptstyle 0.1}$ & $24.7 \pm {\scriptstyle 0.6}$ & $21.5 \pm {\scriptstyle 0.5}$ & \\
         & MinRed & $b=b_r$ & $22.6 \pm {\scriptstyle 0.4}$ & $20.7 \pm {\scriptstyle 0.2}$ & $22.3 \pm {\scriptstyle 0.8}$ & $19.6 \pm {\scriptstyle 0.2}$ & \\
         & \textbf{CLA-R} & $b=b_s+b_r$ & $20.8 \pm {\scriptstyle 0.8}$ & $20.8 \pm {\scriptstyle 0.3}$ & \underline{$26.2 \pm {\scriptstyle 0.8}$} & $\underline{22.3 \pm {\scriptstyle 0.2}}$ & \\
         & \textbf{CLA-E} & $b=b_s+b_r$ &$ \mathbf{24.4 \pm {\scriptstyle 1.3}}$ & $\mathbf{22.5 \pm {\scriptstyle 0.3}}$ & $\mathbf{27.3 \pm {\scriptstyle 1.0}}$ & $\mathbf{23.4 \pm {\scriptstyle 0.3}}$ & \\
         \cmidrule{1-7}
         \multicolumn{2}{c}{SCALE} & $b=b_s+b_r$ & $22.9 \pm {\scriptstyle 0.2}$ & $20.4 \pm {\scriptstyle 0.1}$ & $22.3 \pm {\scriptstyle 1.2}$ & $20.6 \pm {\scriptstyle 0.3}$ & \\
         \multicolumn{2}{c}{Osiris-R} & $b=b_s+b_r$ & $\underline{24.1 \pm {\scriptstyle 0.4}}$  & $\underline{21.9 \pm {\scriptstyle 0.3}}$ & $ 23.5 \pm {\scriptstyle 0.7}$& $19.7 \pm {\scriptstyle 0.6}$ & \\
         \midrule
         \multicolumn{2}{c}{\textit{random encoder}} & - & $15.4 \pm {\scriptstyle 0.4}$ & - & $14.7 \pm {\scriptstyle 0.7}$ & - & \\
         \bottomrule
    \end{tabular}
    \begin{tablenotes}
    \item[1] \footnotesize{LUMP uses interpolation between stream and replay samples.}
    \end{tablenotes}
    \end{threeparttable}
\end{table}
\paragraph{\textsc{Compared Strategies}}
We choose to compare our methods to existing CSSL strategies that are applicable in an online setting, i.e. they do not necessitate task boundaries. Those are: \textit{MinRed}, \textit{Osiris-R}, \textit{SCALE}, \textit{LUMP}.
We also add some baseline strategies that are applicable in this scenario: Experience Replay (ER) \citep{chaudhry2019er_online, buzzega2020rethinking} with a reservoir buffer, ER with a FIFO buffer and simple finetuning.
Note that Osiris-R and SCALE are not paired with SimSiam, as the first one is crafted to work only with contrastive Self-Supervised models, while the second includes a novel SSL loss by itself. We employ a reservoir buffer for SCALE instead of the PSA updated buffer, as the latter would require a forward pass of the entire buffer at each training step, thus breaking our assumptions of computational limitations for OCSSL. 

For comparison purposes, we also include CaSSLe and \textit{CaSSLe-R} as baselines. CaSSLe-R is a simple extension of CaSSLe that includes rehearsal, with the alignment process happening only on replay sample features ($z_{r1}, z_{r2}$), in a similar fashion to CLA. 
Even though CaSSLE and CaSSLe-R are not applicable in an OCSSL setting due to their requirement of task boundaries, we test them nonetheless, providing them with the task boundaries that should otherwise be hidden in this scenario. 
They are included as baselines against which CLA can be compared to observe the impact when the frozen network is substituted with a different alignment target.

\section{Results}
\label{sec:results}
In the \textit{High CBP} configuration (Tab.~\ref{tab:results-high-cbp}),  CLA-E surpasses all other methods in \textsc{Final Acc.}.
The fast adaptation properties of CLA are more evident in ImageNet100, especially with regard to \textsc{Avg. Acc.}, with CLA variants outperforming other methods by at least 3\%. This shows that CLA maintains high accuracy during all the training on the stream.
Specifically, CLA-E is the best-performing method for \textsc{Final Acc.}, while CLA-R shows a better \textsc{Avg. Acc.}, proving that using replay features as targets induces better performance across the entire stream.
\begin{figure}[h]
    \centering
     \begin{subfigure}{0.49\linewidth}
      \centering
      \includegraphics[width=1.0\linewidth]{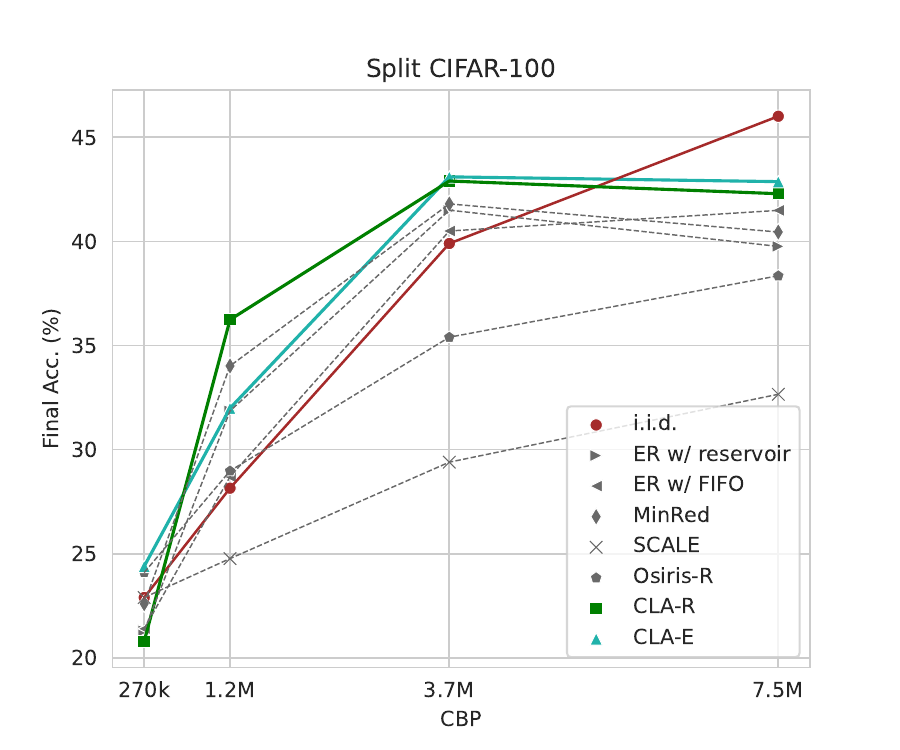}
    \end{subfigure}
    \begin{subfigure}{0.49\linewidth}
      \centering
      \includegraphics[width=1.0\linewidth]{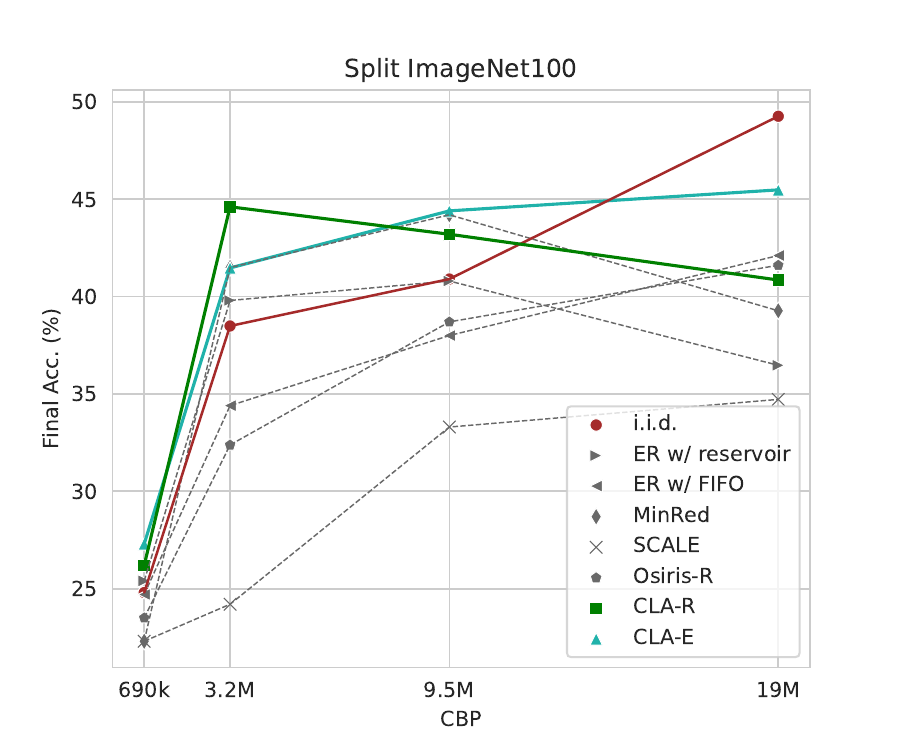}
    \end{subfigure}
    \caption{\textsc{Final Acc.} on Split CIFAR-100 and Imagenet100 for different CBP values. i.i.d. performances increase proportionally to CBP, while on the CL stream performances stagnate after a certain $n_p$.}
    \label{fig:multiple_np_simsiam}
\end{figure}
The additional alignment loss term in CLA ensures fast adaptation, by enforcing current representations to be closer to past ones. It also induces more stability and prevents forgetting of past representations.
Simple rehearsal baselines (\textit{ER w/ FIFO} and \textit{ER w/ reservoir}) achieve competitive scores, surpassing other strategies like Osiris-R and SCALE, which seem to underperform in the \textit{High CBP} setting.
\begin{wraptable}{r}{0.5\textwidth}
    \caption{Ablation experiments comparing different buffers for CLA-R and CLA-E. We report \textsc{Final Acc.} .}
    \label{tab:buffer_ablation}
    \centering
    \begin{tabular}{cccc}
    \toprule
         & \textsc{Buffer Type} & CIFAR100 & ImageNet100 \\
         \cmidrule(rl){2-2}\cmidrule(rl){3-3} \cmidrule(rl){4-4}
         \multirow{3}{*}{CLA-R}
         & reservoir & $40.7 \pm {\scriptstyle 1.1}$ & $36.5 \pm {\scriptstyle 0.5}$  \\
         &  MinRed  & $40.8 \pm {\scriptstyle 0.4}$ &  $40.6 \pm {\scriptstyle 0.6}$ \\
         &  \textbf{FIFO} & $42.9 \pm {\scriptstyle 0.3}$ & $43.2 \pm {\scriptstyle 0.4}$  \\
         \midrule
         \multirow{3}{*}{CLA-E} &  reservoir & $40.1 \pm {\scriptstyle 0.3}$ &  $34.4 \pm {\scriptstyle 0.3}$\\
         &  MinRed & $41.2 \pm {\scriptstyle 0.3}$ & $ 37.6 \pm {\scriptstyle 0.4}$ \\
         &  \textbf{FIFO} & $43.1 \pm {\scriptstyle 0.3}$ & $44.4 \pm {\scriptstyle 0.9}$  \\
         \bottomrule
    \end{tabular}
\end{wraptable}
The \textit{Low CBP} scenario (Tab. \ref{tab:results-low-cbp}), as expected, shows lower overall accuracy scores due to the lower computational budget.
More importantly, when comparing strategies using $n_p=3$ and no additional replay samples against strategies using  $n_p=1$ and extending the minibatch to $b=30$, we observe comparable accuracies, thus proving the validity of our CBP metric.
In \textit{Low CBP}, again, CLA-E surpasses other strategies in both datasets, with CLA-R being comparable to CLA-E. Even CLA-b has good performances surpassing all other strategies with limited $b=10$; simple replay strategies are again a strong baseline.
Unlike \textit{High CBP}, in \textit{Low CBP} scenario both Osiris-R and SCALE achieve competitive scores.
Moreover, CLA is capable to surpass the CaSSLe and CaSSLe-R baselines, which we recall are not applicable in OCSSL, in the \textit{High CBP} setting, while reaching similar performance in \textit{Low CBP}. This proves that having a target for alignment that does not require task boundaries does not hurt performance, but instead, using EMA network and replayed features as alignment target has the potential of also improving offline CSSL strategies.   

A surprising result regards the i.i.d. training: it is commonly considered in CL as the upper bound, but in our scenario i.i.d. is often reached or surpassed by various OCSSL strategies.
In particular, CLA is able to match or surpass its i.i.d. baselines in all cases.
This phenomenon is especially present when comparing SimSiam based strategies, with CLA-E surpassing i.i.d. in \textsc{Final Acc.} by 3\%, in the \textit{High CBP} setting.
This hints towards i.i.d. training not being the most efficient training paradigm for \emph{early} stages of the training process and with a limited computational budget.
Our hypothesis is that this phenomenon is caused by two factors: 1) the presence of replay, as more training iterations are performed on buffer samples; and 2) the fast adaptation properties induced by alignment in CLA.

\paragraph{\textsc{Buffer Ablation}}
We conducted an ablation study trying different buffers for CLA-E and CLA-R. We compared reservoir, MinRed and FIFO buffers. Results in Table~\ref{tab:buffer_ablation} confirm that the best buffer choice for both CLA-E and CLA-R is FIFO.

\paragraph{\textsc{Changing $n_p$}}
\begin{wrapfigure}{r}{0.59\textwidth}
    \centering
    \includegraphics[width=1.0\linewidth]{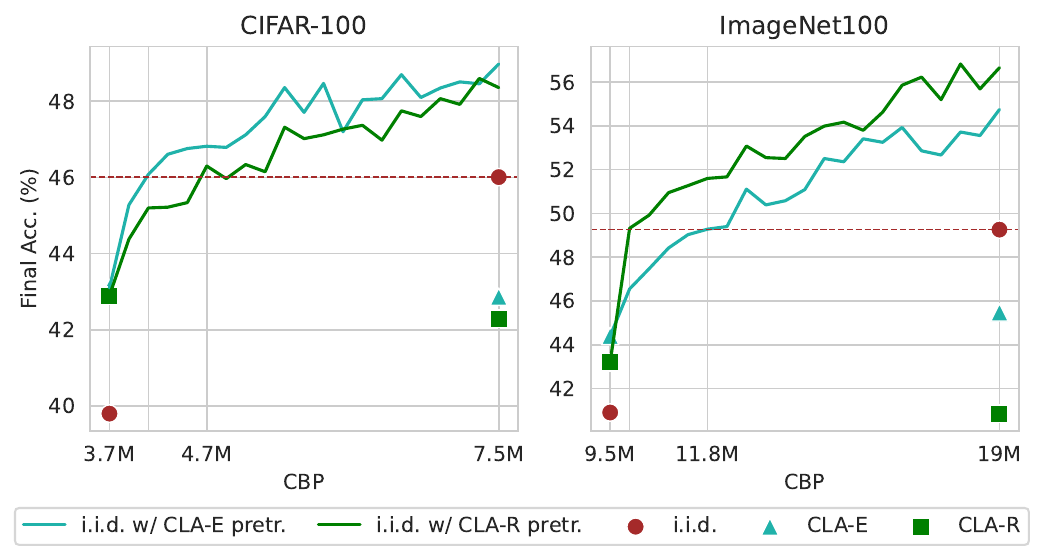}
    \caption{Continuing the pretraining with i.i.d. data of a CLA-pretrained learner, yields better results than a full i.i.d. training. This proves that CLA is more efficient than i.i.d. in early stages of SSL training without impacting future training performance. }
    \label{fig:pretraining-simsiam}
\end{wrapfigure}
In Figure~\ref{fig:multiple_np_simsiam} we compare the probing \textsc{Final Acc.} of strategies with different CBP budgets. 
We include the results for the \textit{High} and \textit{Low CBP} scenarios and also include additional experiments with the same settings of \textit{High CBP}, but with different number of minibatch passes, $n_p=1$ and $n_p=6$, instead of $n_p=3$.
The main observation is that increasing $n_p$ after a certain amount does not increase the performance of the model. This indicates that a limit of current models is not being able to exploit higher CBP if $b$ or the buffer size are not increased as well.
More research is needed to design online methods that scale to larger computational budgets without the need for more replay samples.
Of course, this is not the case for i.i.d., as more training budget correspondingly increases the performance, leading it to surpass all strategies trained on the OCSSL stream for $n_p=6$.
In addition, in most cases, the FIFO buffer behaves better in the highest CBP settings, whereas strategies with buffers that change less frequently (MinRed and reservoir) converge faster with low CBP. However, they tend to overfit and do not exploit high CBP as much.

\paragraph{\textsc{Efficient Early Pretraining with CLA}}
Our results hint at the fact that CLA is able to speed up the early phases of training with respect to i.i.d. We hypothesize that, during the initial phases of Self-Supervised training (i.e. with a low CBP), CLA can learn high quality representations even with a small CBP budget.
To this end, we experimented by using a CLA pretrained learner (checkpoints from \textit{High CBP} experiments) that is subsequently trained with i.i.d. We compared its performance with a fully i.i.d. trained network.
The learning rate for the full i.i.d. training was tuned using the same hyperparameter selection process as the main experiments. The learning rate found in this way was also used to train the i.i.d. portion of i.i.d. with CLA pretraining.
Results (Fig. \ref{fig:pretraining-simsiam}) show that we can achieve the same level of performance of full i.i.d. training with just a short i.i.d. training on a CLA pretrained Self-Supervised learner.
\begin{wraptable}{r}{0.5\textwidth}
    \centering
    \caption{CIFAR-100 pretrained encoders evaluated on SVHN and Imagenet100 pretrained encoders evaluated on Stanford Cars. Best in \textbf{bold}, second best \underline{underlined}.}
    \label{tab:downstream}
    \begin{tabular}{ccc}
    \toprule
        \textsc{Method} & CIFAR $\xrightarrow{}$ SVHN & ImageNet $\xrightarrow{}$ Cars \\
         
         \cmidrule(rl){1-1} \cmidrule(rl){2-2} \cmidrule(rl){3-3}
         \textit{i.i.d.} & $59.8 \pm {\scriptstyle 0.1}$ & $9.2 \pm {\scriptstyle 0.7}$  \\
         ER w/ reservoir & $59.8 \pm {\scriptstyle 0.1}$ & $10.0 \pm {\scriptstyle 1.0}$  \\
         ER w/ FIFO & $58.9 \pm {\scriptstyle 0.1}$ & $8.3 \pm {\scriptstyle 0.8}$  \\
         MinRed & $58.2 \pm {\scriptstyle 0.1}$ & $13.3 \pm {\scriptstyle 0.1}$  \\
        SCALE & $47.8 \pm {\scriptstyle 0.1}$ & $6.0 \pm {\scriptstyle 1.1}$  \\
         Osiris-R & $\underline{60.7 \pm {\scriptstyle 0.3}}$ & $7.7 \pm {\scriptstyle0.4}$  \\
         \midrule
         \textbf{CLA-R} & $\mathbf{65.1 \pm {\scriptstyle0.0}}$ & $\underline{15.5 \pm {\scriptstyle 0.1}}$  \\
         \textbf{CLA-E} & $58.3 \pm {\scriptstyle0.0}$ & $\mathbf{15.8 \pm {\scriptstyle 0.4}}$  \\
         \bottomrule
    \end{tabular}
\end{wraptable}
This suggests that CL strategies that favor fast adaptation can be used with excellent results in i.i.d. scenarios, with a large advantage in training efficiency.
Moreover, this also highlights that CLA is not only able to quickly learn new features, but it can also serve as a strong initialization for further training.

\paragraph{\textsc{Downstream Task}}
We also evaluated the models pretrained in the \textit{High CBP} scenario on downstream tasks that are different from the original pretraining dataset.
We chose to evaluate by probing on Stanford Cars \citep{krause2013cars} dataset using ImageNet100 pretrained encoders, and on SVHN \citep{netzer2011svhn} using CIFAR100 pretrained encoders. Results are reported in Table \ref{tab:downstream}.
CLA-E obtains the highest probing accuracy on SVHN, while on Cars both CLA-E and CLA-R distinctly surpass all other methods.

\section{Conclusion}
In this work, we introduced CLA, a novel strategy tailored for the OCSSL scenario. CLA mitigates forgetting by aligning current model representations with past representations without the need for task boundaries, enabling faster convergence and robust performance across non-stationary data streams. 
Our experiments showed that CLA surpasses existing methods, and even i.i.d. training, under a constrained computational budget, defined by our CBP metric.
Additionally, we observed that CLA, when used as an early-stage pretraining protocol, improves final model performance over full i.i.d. pretraining.

The OCSSL scenario is still largely understudied. However, designing more effective OCSSL approaches can have a profound impact on CL and even on deep learning as a whole. In fact, learning how to quickly build robust representations from unlabeled data is a fundamental objective for many deep learning models, like large foundation models. Reducing the time (hence, the cost) of the pretraining phase of a foundation model without reducing its downstream performance is key to enable sustainable developments of deep learning solutions.

\section*{Acknowledgements}
This paper has been partially supported by the CoEvolution project, funded by EU Horizon 2020 under GA n 101168559. We acknowledge projects PID2022-143257NB-I00, financed by MCIN/AEI/10.13039/501100011033 and FSE+. We acknowledge ISCRA for awarding this project access to the LEONARDO supercomputer, owned by the EuroHPC Joint Undertaking, hosted by CINECA (Italy).

\bibliography{mybib}
\bibliographystyle{collas2025_conference}

\appendix
\section{CLA pseudocode}
\label{sec:pseudocode}

\begin{algorithm}[h]
\caption{Pseudocode for \textbf{CLA-b} training loop}
\label{alg:cla-b}
\begin{algorithmic}[1]
    \For{$x$ in $\mathcal{D}$}
    \For{ $p$ in $n_p$}
        \State $x_1, \ x_2$ $\gets$ Augmentations($x$)
        \State $z_1, \ z_2 \ \gets \ \theta(x_1),\ \theta(x_2)$
        \State $\theta' \ \leftarrow \ \tau \theta' + (1-\tau)\theta $ \Comment{EMA network update}
        \State $z_1', \ z_2' \ \gets \ \theta'(x_1, \ x_2)$ \Comment{Ema network forward pass}
        \State $\mathcal{L}_\textit{reg\_CLA-b} \gets \frac{-S_C(a_\phi (z_1), z_1')}{2} +
    \frac{-S_C(a_\phi(z_2), z_2')}{2}$ \Comment{CLA-b alignment loss}

        \State $\mathcal{L} \ \gets \ \mathcal{L}_\textit{SSL}(z_1, z_2) + \omega   \mathcal{L}_\textit{reg\_CLA-b}$
        \State Backprop($\mathcal{L}$)
    \EndFor
    \EndFor
\end{algorithmic}
\end{algorithm}

\begin{algorithm}
\caption{Pseudocode for \textbf{CLA-E/R} training loop}
\label{alg:cla}
\begin{algorithmic}[1]
    \For{$x$ in $\mathcal{D}$}
    \For{ $p$ in $n_p$}
        \State $x_r, z^* \sim \mathcal{M}$      \Comment{Sampling from buffer $\mathcal{M}$}
        \State $x_1 \cup x_{r1}, \ x_2 \cup x_{r2}$ $\gets$ Augmentations($x \cup x_r$)
        \State $z_1 \cup z_{r1}, \ z_2 \cup z_{r2} \ \gets \ \theta(x_1 \cup x_{r1}),\ \theta(x_2 \cup x_{r2} )$
        \If{CLA-E}
            \State $\theta' \ \leftarrow \ \tau \theta' + (1-\tau)\theta $ \Comment{EMA network update}
            \State $\hat{z}_{r1}, \ \hat{z}_{r2} \ \gets \ \theta'(x_{r1}, \ x_{r2})$  \Comment{Ema network forward pass}            
            \State  $\mathcal{L}_\textit{reg\_CLA-E} \gets \frac{-S_C(a_\phi (z_{r1}), \hat{z}_1)}{2} + \frac{-S_C(a_\phi(z_{r2}), \hat{z}_2)}{2}$ \Comment{CLA-E alignment loss}
        \ElsIf{CLA-R}
            \State $\mathcal{L}_\textit{reg\_CLA-R} \gets \frac{-S_C(a_\phi (z_{r1}), z^*)}{2} + \frac{-S_C(a_\phi(z_{r2}), z^*)}{2}$ \Comment{CLA-R alignment loss}
        \EndIf

        \State $\mathcal{L} \ \gets \ \mathcal{L}_\textit{SSL}(z_1 \cup z_{r1}, z_2 \cup z_{r2}) + \omega  \mathcal{L}_\textit{reg\_CLA-E/R}$

        \State Backprop($\mathcal{L}$)

    \EndFor
    \State Update $\mathcal{M}$ with new samples and features $(x, \frac{z_1 + z_2}{2})$
    \State Update old features of $x_r$ in $\mathcal{M}$: 
        \State \ \ \ \ \ \ \ \  $z^* \gets 0.5 \cdot z^* + 0.5 \cdot \frac{z_{r1}+z_{r2}}{2}$
    \EndFor
\end{algorithmic}
\end{algorithm}

In this Section we present the pseudocode for CLA strategy. Alg. \ref{alg:cla-b} shows the training loop of an SSL method training on an OCSSL stream $\mathcal{D}$ paired with CLA-b strategy. 
Instead, Alg. \ref{alg:cla} illustrates the same training loop but with CLA-E/R strategy. 
Notable differences between the two are the inclusion of the memory buffer $\mathcal{M}$ in CLA-E/R, with memory sampling and memory update steps, and, consequently, the alignment process only being applied to replayed samples features $z_r1, z_r2$.

\section{SimCLR experiments}
\label{sec:simclr-experiments}

\begin{table*}[h]
    \centering
    \caption{\textbf{Results for High CBP} with \textbf{SimCLR} on Split-CIFAR100 and Split-Imagenet100, with CBP = 3.7M for CIFAR-100 and CBP = 9.5M for ImageNet100, using $b_s=10$, $b=138$, $n_p=3$. Best in \textbf{bold}, second best \underline{underlined}.}
    \label{tab:results-high-cbp-simclr}
    \begin{tabular}{cccccc}
    \toprule
          & & \multicolumn{2}{c}{CIFAR100 (20 exps)} &
         \multicolumn{2}{c}{ImageNet100 (20 exps)} \\
         \multicolumn{2}{c}{\textsc{Method}} & \textsc{Final Acc.} & \textsc{Avg. Acc.} & \textsc{Final Acc.} & \textsc{Avg. Acc.} \\
         \cmidrule(rl){1-2}\cmidrule(rl){3-4} \cmidrule(rl){5-6}
         \multirow{7}{*}{\rotatebox[origin=c]{90}{\textsc{SimCLR}}} 
         & \textit{i.i.d.} & $42.9 \pm {\scriptstyle 0.3}$ & - & $48.5 \pm {\scriptstyle 0.7}$ & - \\
         & \textit{CaSSLe-R} & $40.1 \pm {\scriptstyle 0.5}$ & $36.5 \pm {\scriptstyle 0.2}$ & $40.6 \pm {\scriptstyle 0.2}$ & $37.4 \pm {\scriptstyle 0.2}$ \\
         & ER w/ reservoir & $40.9 \pm {\scriptstyle 0.2}$ & $36.8 \pm {\scriptstyle 0.2}$ & $39.2 \pm {\scriptstyle 0.2}$ & $33.5 \pm {\scriptstyle 0.2}$  \\
         & ER w/ FIFO & $41.8 \pm {\scriptstyle 0.2}$ & $36.6 \pm {\scriptstyle 0.2}$ & $38.1 \pm {\scriptstyle 0.5}$ & $32.3 \pm {\scriptstyle 0.2}$  \\
         & MinRed & $41.7 \pm {\scriptstyle 0.2}$ & $\underline{37.9 \pm {\scriptstyle 0.1}}$ & $41.3 \pm {\scriptstyle 0.6}$ & $35.1 \pm {\scriptstyle 0.3}$  \\
         & \textbf{CLA-R} & $\mathbf{42.6 \pm {\scriptstyle 0.3}}$ & $\mathbf{38.4 \pm {\scriptstyle 0.1}}$ & $\underline{44.2 \pm {\scriptstyle 0.4}}$ & $\mathbf{39.7 \pm {\scriptstyle 0.0}}$  \\
         & \textbf{CLA-E} & $\underline{42.0 \pm {\scriptstyle 0.3}}$ & $35.9 \pm {\scriptstyle 0.3}$ & $\mathbf{45.1 \pm {\scriptstyle 0.2}}$ & $\underline{38.3 \pm {\scriptstyle 0.3}}$  \\
         \midrule
          \multicolumn{2}{c}{SCALE} & $29.4 \pm {\scriptstyle 0.7}$ & $26.2 \pm {\scriptstyle 0.2}$ & $33.3 \pm {\scriptstyle 0.6}$ & $28.6 \pm {\scriptstyle0.5}$ \\
         \multicolumn{2}{c}{Osiris-R} & $35.4 \pm {\scriptstyle 0.2}$ & $30.4 \pm {\scriptstyle 0.2}$ & $38.7 \pm {\scriptstyle 0.2}$ & $32.8 \pm {\scriptstyle 0.2}$  \\
         \multicolumn{2}{c}{\textit{random encoder}} & $15.4 \pm {\scriptstyle 0.4}$ & - & $14.7 \pm {\scriptstyle 0.7}$ & - \\
         \bottomrule
    \end{tabular}
\end{table*}

\begin{table*}[h]
    \centering
    \caption{\textbf{Results for Low CBP} with \textbf{SimCLR} on Split-CIFAR100 and Split-Imagenet100, with CBP = 270k for CIFAR-100 and CBP =  for ImageNet100, using $b_s=10$, $b=30$ and $n_p=1$ for strategies using minibatch with additional replay samples, while $b=10$ and $n_p=3$ for fixed size minibatch strategies.}
    \label{tab:results-low-cbp-simclr}
    \begin{tabular}{ccccccc}
    \toprule
        & & \multicolumn{2}{c}{CIFAR100 (20 exps)} &  
         \multicolumn{2}{c}{ImageNet100 (20 exps)} & \\
          \multicolumn{2}{c}{\textsc{Method}} & \textsc{Final Acc.} & \textsc{Avg. Acc.} & \textsc{Final Acc.} & \textsc{Avg. Acc.} & \\
          \cmidrule(rl){1-2}\cmidrule(rl){3-4} \cmidrule(rl){5-6}
         
         \multirow{13}{*}{\rotatebox[origin=c]{90}{\textsc{SimCLR}}} 
         & \textit{i.i.d.} & $25.8 \pm {\scriptstyle 0.3}$ & - & $24.2 \pm {\scriptstyle 1.1}$ & - &  
          \multirow{5}{*}{\rotatebox[origin=c]{90}{\footnotesize\textsc{$b=10$, $n_p=3$}}}\\
         & \textit{CaSSLe} & $25.1 \pm {\scriptstyle 0.6}$ & $23.8 \pm {\scriptstyle 0.1}$ & $26.3 \pm {\scriptstyle 0.6}$ & $23.0 \pm {\scriptstyle 0.1}$ & \\
         & finetuning & $25.2 \pm {\scriptstyle 0.3}$ & $23.5 \pm {\scriptstyle 0.1}$ & $25.2 \pm {\scriptstyle 0.4}$ & $22.6 \pm {\scriptstyle 0.1}$ & \\
         & LUMP & $25.6 \pm {\scriptstyle 0.2}$ & $23.4 \pm {\scriptstyle 0.2}$ & $25.3 \pm {\scriptstyle 1.2}$ & $23.2 \pm {\scriptstyle 0.4}$  & \\
         & \textbf{CLA-b} & $25.6 \pm {\scriptstyle 0.2}$ & $23.5 \pm {\scriptstyle 0.1}$ & $26.3 \pm {\scriptstyle 1.3}$ & $22.9 \pm {\scriptstyle 0.1}$  & \\  
         \cmidrule{2-6}
         & \textit{i.i.d.} & $25.6 \pm {\scriptstyle 0.3}$ & - & $30.1 \pm {\scriptstyle 1.7}$ & - & \multirow{9}{*}{\rotatebox[origin=c]{90}{\footnotesize\textsc{$b=30$, $n_p=1$}}} \\
         & \textit{CaSSLe-R} & $27.0 \pm {\scriptstyle 0.1}$ & $24.2 \pm {\scriptstyle 0.2}$ & $31.7 \pm {\scriptstyle 0.2}$ & $26.1 \pm {\scriptstyle 0.3}$  & \\
         & ER w/ reservoir & $24.5 \pm {\scriptstyle 0.2}$ & $22.4 \pm {\scriptstyle 0.1}$ & $27.4 \pm {\scriptstyle 1.0}$ & $24.2 \pm {\scriptstyle 0.1}$  & \\
         & ER w/ FIFO & $24.4 \pm {\scriptstyle 0.2}$ & $22.2 \pm {\scriptstyle 0.1}$ & $27.5 \pm {\scriptstyle 0.2}$ & $24.1 \pm {\scriptstyle 0.1}$ & \\
         & MinRed & $25.9 \pm {\scriptstyle 0.4}$ & $22.8 \pm {\scriptstyle 0.1}$ & $29.6 \pm {\scriptstyle 0.4}$ & $\underline{25.8 \pm {\scriptstyle 0.1}}$ & \\
         & \textbf{CLA-R} & $\underline{26.5 \pm {\scriptstyle 0.2}}$ & $\mathbf{24.1 \pm {\scriptstyle 0.1}}$ & $\mathbf{30.9 \pm {\scriptstyle 0.9}}$ & $\mathbf{26.1 \pm {\scriptstyle 0.5}}$ & \\
         & \textbf{CLA-E} & $\mathbf{26.7 \pm {\scriptstyle 0.4}}$ & $\mathbf{24.1 \pm {\scriptstyle 0.0}}$ & $\underline{30.6 \pm {\scriptstyle 0.8}}$ & $25.3 \pm {\scriptstyle 0.1}$ & \\
         \cmidrule{1-6}
         \multicolumn{2}{c}{SCALE} & $22.9 \pm {\scriptstyle 0.2}$ & $20.4 \pm {\scriptstyle 0.1}$ & $22.3 \pm {\scriptstyle 1.2}$ & $20.6 \pm {\scriptstyle 0.3}$ & \\
        \multicolumn{2}{c}{Osiris-R} & $24.1 \pm {\scriptstyle 0.4}$  & $21.9 \pm {\scriptstyle 0.3}$ & $23.5 \pm {\scriptstyle 0.7}$& $19.7 \pm {\scriptstyle 0.6}$ & \\
        \midrule
        \multicolumn{2}{c}{\textit{random encoder}} & $15.4 \pm {\scriptstyle 0.4}$ & - & $14.7 \pm {\scriptstyle 0.7}$ & - & \\
         \bottomrule
    \end{tabular}

\end{table*}

In this section, we replicate the experiments from Sections \ref{sec:experiments} and \ref{sec:results} using SimCLR \citep{chen2020simclr} instead of SimSiam. The primary motivation for this choice is to evaluate our findings on an additional self-supervised learning (SSL) model, thereby testing the proposed and existing strategies with a different type of SSL loss, specifically the contrastive loss employed by SimCLR.
We only execute the additional experiments on strategies that need an underlying Self-Supervised learner, leaving the other strategies (i.e. SCALE, Osiris-R) unchanged. We adhere to the same experimental settings described in Section \ref{sec:experiments}.

\subsection{High CBP}
Table~\ref{tab:results-high-cbp-simclr} presents the results for the \textit{High CBP} setting, defined as CBP = 3.7M for CIFAR-100 and CBP = 9.5M for ImageNet100, with hyperparameters set to $b_s=10$, $b=138$, and $n_p=3$.
Using a SimCLR backbone yields results consistent with those observed for SimSiam, where CLA achieves the best performance across both benchmarks. A notable distinction is the larger accuracy gap achieved by CLA on ImageNet100 compared to other methods, while on CIFAR-100, CLA-E scores are very close to those of approaches that omit loss regularization terms, such as ER and MinRed.
It is also worth noting that SimCLR demonstrates higher i.i.d. baselines than SimSiam, achieving results, remaining unbeaten by methods trained on the OCSSL stream. We hypothesize that this is due to SimCLR’s inherently faster convergence compared to SimSiam.

\subsection{Low CBP}
Table~\ref{tab:results-low-cbp-simclr} presents the results for the \textit{Low CBP} setting. Accuracies with a SimCLR backbone are higher than those obtained with SimSiam, supporting our hypothesis that SimCLR achieves faster convergence under limited computational budgets, particularly during the early stages of pretraining.
CLA achieves the highest scores across both metrics, although MinRed performs competitively, with accuracy levels closely matching those of CLA. Notably, methods employing $n_p=3$ and $b=10$ on ImageNet100 exhibit lower overall scores than $n_p=1$ and $b=30$, highlighting the influence of minibatch size as a limiting factor for contrastive SSL performance. This aligns with findings from prior studies \citep{chen2020simclr, he2020moco}, which demonstrate that increasing the number of negative samples enhances representation quality. These results underscore the necessity of extending minibatch sizes with replay samples in an OCSSL scenario, as small minibatch sizes can significantly impair the performance of certain SSL methods.

\subsection{Changing $\mathbf{n_p}$}
\begin{figure}[t]
    \centering
     \begin{subfigure}{0.49\linewidth}
      \centering
      \includegraphics[width=1.0\linewidth]{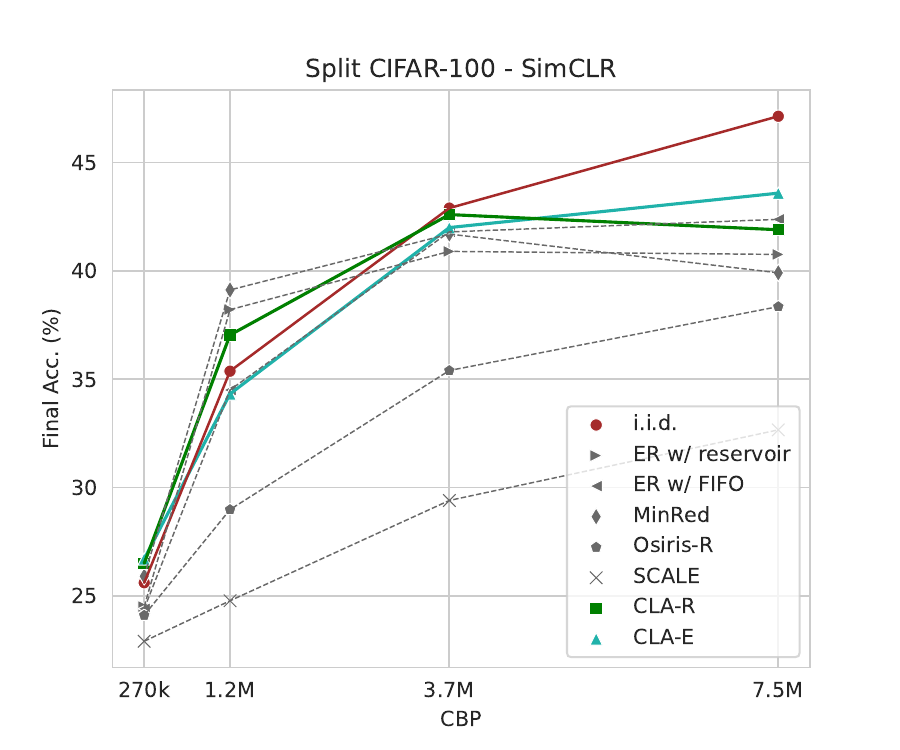}
    \end{subfigure}
    \begin{subfigure}{0.49\linewidth}
      \centering
      \includegraphics[width=1.0\linewidth]{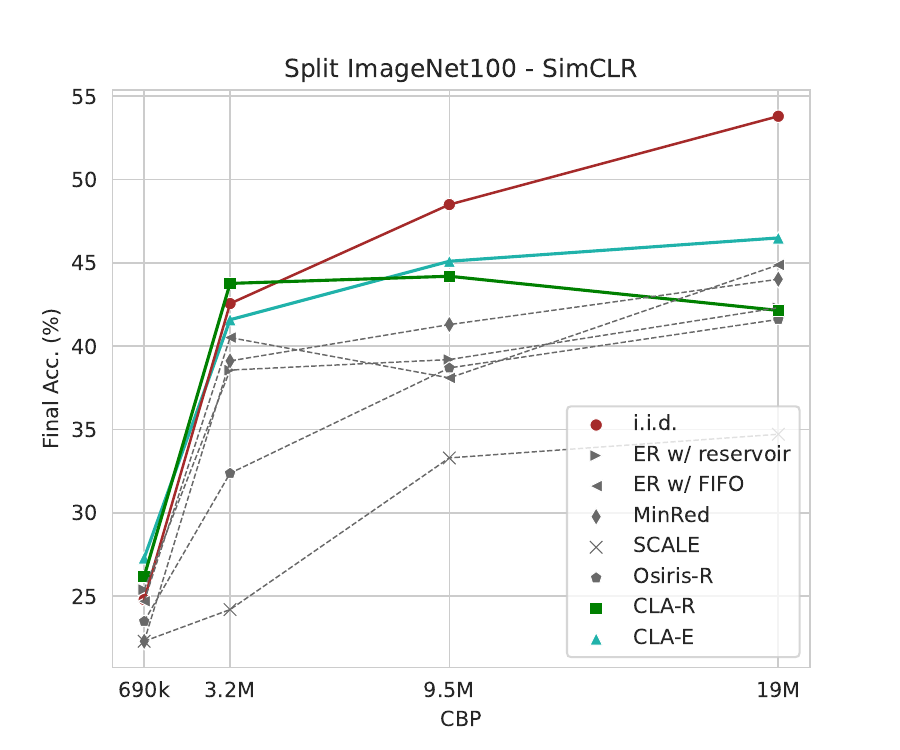}
    \end{subfigure}
    \caption{\textsc{Final Acc.} on Split CIFAR-100 and Imagenet100 for different CBP values. i.i.d. performances increase proportionally to CBP, while on the CL stream performances stagnate after a certain $n_p$.}
    \label{fig:multiple_np_simclr}
\end{figure}
Figure~\ref{fig:multiple_np_simclr} illustrates the results for strategies incorporating replay samples into the minibatch, evaluated across different CBP training budgets. We report both \textit{Low CBP} and \textit{High CBP} results, along with additional experiments conducted in the \textit{High CBP} setting using $n_p=1$ and $n_p=6$ (instead of $n_p=3$) to explore varying CBP values. All methods requiring an SSL backbone (excluding Osiris-R and SCALE) use SimCLR.
The trends closely mirror those observed with SimSiam: i.i.d. accuracy scales proportionally with the training budget, whereas performance on the continual learning (CL) stream plateaus beyond a certain $n_p$. Notably, in the CIFAR-100 dataset with CBP = 1.2M, both MinRed and ER with a reservoir buffer outperform CLA, highlighting the advantages of subset-based buffering in low-CBP scenarios.
In the highest CBP setting, CLA-E demonstrates superior performance, indicating its versatility across different computational budgets. In contrast, CLA-R exhibits slightly poorer results at CBP = 19M in ImageNet100, likely due to the regularization from aligning on the same buffer features becoming overly restrictive.

\subsection{Efficient Early Pretraining with CLA}
\begin{figure}[b]
    \centering
    \includegraphics[width=0.6\linewidth]{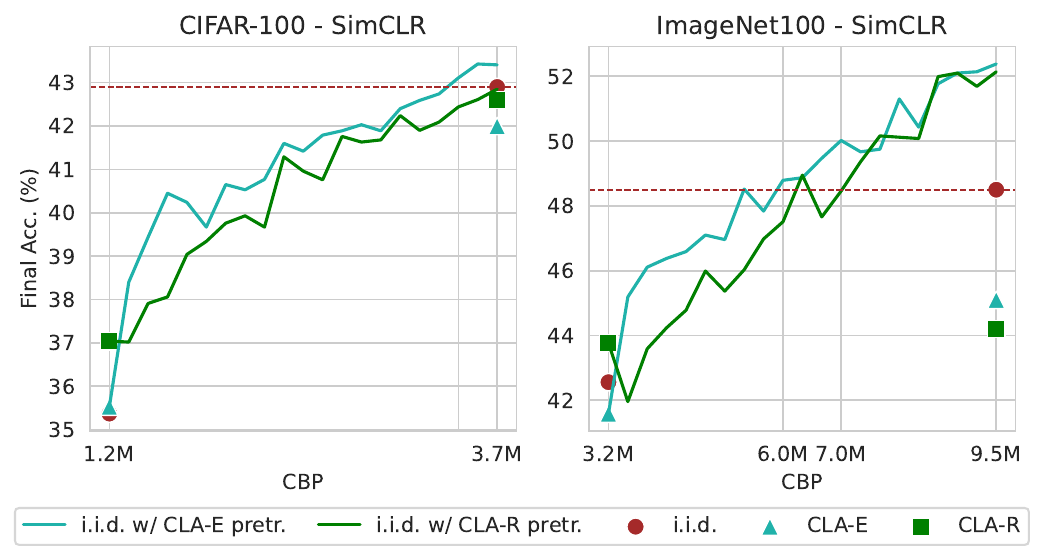}
    \caption{Continuing the pretraining with i.i.d. data of a CLA-pretrained learner (using SimCLR as the SSL backbone), yields better results than a fully i.i.d. trained model, on ImageNet100. On CIFAR-100, instead, we are able to only reach i.i.d. performances.}
    \label{fig:pretraining-simclr}
\end{figure}
In Figure~\ref{fig:pretraining-simclr}, we conduct experiments using a CLA-pretrained learner, subsequently trained in an i.i.d. setting, but using SimCLR instead of SimSiam backbone. We compare its performance to that of a network fully trained in an i.i.d. manner, maintaining an equivalent CBP budget. Unlike the SimSiam experiments, we use lower initial CBP values of 1.2M for CIFAR-100 and 3.2M for ImageNet100 during CLA pretraining, targeting final CBP values of 3.7M for CIFAR-100 and 9.2M for ImageNet100. This choice reflects SimCLR generally faster convergence (as shown in Tab.~\ref{tab:results-low-cbp-simclr}), making the benefits of efficient early pretraining apparent at earlier stages of self-supervised training.
The results on ImageNet100 confirm that CLA's rapid adaptation capabilities can enhance performance even in an i.i.d. scenario. In contrast, on CIFAR-100, CLA pretraining does not significantly outperform full i.i.d. training, suggesting that these fast adaptation properties are more impactful for complex datasets when using a contrastive SSL method. This observation is consistent with Tab.~\ref{tab:results-high-cbp-simclr}, where CLA shows a more pronounced performance advantage over other strategies on ImageNet100 compared to CIFAR-100.

\subsection{Downstream Tasks}
\begin{table}[t]
    \centering
    \caption{Results for probing on a different downstream task. CIFAR-100 pretrained encoders evaluated on SVHN and Imagenet100 pretrained encoders evaluated on Stanford Cars. \textbf{SimCLR} backbones were used paired with strategies. Best in \textbf{bold}, second best \underline{underlined}.}
    \label{tab:downstream-simclr}
    \begin{tabular}{ccc}
    \toprule
        \textsc{Method} & CIFAR $\xrightarrow{}$ SVHN & ImageNet $\xrightarrow{}$ Cars \\
         
         \cmidrule(rl){1-1} \cmidrule(rl){2-2} \cmidrule(rl){3-3}
         \textit{i.i.d.} & $59.1 \pm {\scriptstyle 0.2}$ & $11.4 \pm {\scriptstyle 0.4}$  \\
         ER w/ reservoir & $59.0 \pm {\scriptstyle 0.1}$ & $8.5 \pm {\scriptstyle 0.6}$  \\
         ER w/ FIFO & $59.1 \pm {\scriptstyle 0.2}$ & $6.9 \pm {\scriptstyle 0.6}$  \\
         MinRed & $58.3 \pm {\scriptstyle 0.1}$ & $8.6 \pm {\scriptstyle 0.8}$  \\
        SCALE & $47.8 \pm {\scriptstyle 0.1}$ & $6.0 \pm {\scriptstyle 1.1}$  \\
         Osiris-R & $\underline{60.7 \pm {\scriptstyle 0.3}}$ & $7.7 \pm {\scriptstyle0.4}$  \\
         \midrule
         \textbf{CLA-R} & $\mathbf{61.1 \pm {\scriptstyle0.1}}$ & $\mathbf{13.8 \pm {\scriptstyle 0.6}}$  \\
         \textbf{CLA-E} & $57.4 \pm {\scriptstyle0.2}$ & $\underline{13.3 \pm {\scriptstyle 0.5}}$  \\
         \bottomrule
    \end{tabular}
\end{table}
Following the approach used with SimSiam, we conduct evaluations on downstream tasks that differ from the original pretraining dataset. Specifically, we perform linear probing by testing ImageNet100 Self-Supervised pretrained models on the Stanford Cars dataset and CIFAR-100 Self-Supervised pretrained models on the SVHN dataset.
The results (Tab.~\ref{tab:downstream-simclr}) are consistent with those obtained using SimSiam. On the SVHN dataset, CLA-R achieves the highest score, with the other models exhibiting competitive performance. In contrast, on Stanford Cars, both CLA-E and CLA-R outperform all other models by at least 4\% in accuracy. These results further demonstrate the effectiveness of our proposed methods, particularly when pretrained on a more complex dataset like ImageNet100.

\section{Hyperparameters}
\label{sec:hyperparams}
For all methods, we performed a grid search on a 10\% validation set to identify the optimal learning rate and, for the proposed CLA methods, also for the best regularization strength parameter, $\omega$. 
The best hyperparameter of the grid search was selected using \textsc{Final Accuracy}, obtained by linear probing.
All other hyperparameters of existing methods were set as in their original implementations.

\subsection{Best Hyperparameters}
\label{sec:best-hyperparams}
\begin{table*}[!ht]
    \centering
    \caption{Best hyperparameters for \textit{High CBP} setting on Split-CIFAR100 and Split-Imagenet100.}
    \label{tab:hyperparams_high}
    \begin{tabular}{cccc}
    \toprule
         \multicolumn{2}{c}{\textsc{Method}} & CIFAR100 (20 exps) & 
         ImageNet100 (20 exps) \\
         \cmidrule{1-2} \cmidrule(rl){3-3} \cmidrule(rl){4-4}
         \multirow{7}{*}{\rotatebox[origin=c]{90}{\textsc{SimSiam}}}
         &\textit{i.i.d.} & lr: 0.1 & lr: 0.03\\
         & \textit{CaSSLe-R} & lr: 0.3 & lr: 0.1 \\
         & ER w/ reservoir & lr: 0.1 & lr: 0.03 \\
         & ER w/ FIFO & lr: 0.1 & lr: 0.03 \\
         & MinRed & lr: 0.1 & lr: 0.03 \\
         & CLA-R & lr: 0.3, $\omega$: 1.0 & lr: 0.1, $\omega$: 3.0 \\
         & CLA-E & lr: 0.3, $\omega$: 0.3 & lr: 0.1 , $\omega$: 0.3 \\
         \midrule
         \multirow{8}{*}{\rotatebox[origin=c]{90}{\textsc{SimCLR}}}
         & \textit{i.i.d.} & lr: 0.1 & lr: 0.03 \\
         & \textit{CaSSLe-R} & lr: 0.3 & lr: 0.03 \\
         & ER w/ reservoir & lr: 0.1 & lr: 0.01 \\
         & ER w/ FIFO & 0.1 & lr: 0.01  \\
         & MinRed & 0.1 & lr: 0.01  \\
         & Osiris-R & lr: 0.01 & lr: 0.003  \\
         & CLA-R & lr: 0.3, $\omega$: 3.0 & lr: 0.03, $\omega$: 1.0  \\
         & CLA-E & lr: 0.3, $\omega$: 1.0 & lr: 0.03, $\omega$: 1.0  \\
         \midrule
         \multicolumn{2}{c}{SCALE} & lr: 0.003 & lr: 0.003 \\
         \bottomrule
    \end{tabular}
    
\end{table*}

\begin{table*}[!ht]
    \centering
    \caption{Best hyperparameters for \textit{Low CBP} setting on Split-CIFAR100 and Split-Imagenet100.}
    \label{tab:hyperparams_low}
    \begin{tabular}{ccccc}
    \toprule
         \multicolumn{2}{c}{\textsc{Method}} & CIFAR100 (20 exps) & 
         ImageNet100 (20 exps) &\\
         \cmidrule{1-2} \cmidrule(rl){3-3} \cmidrule(rl){4-4}
         \multirow{12}{*}{\rotatebox[origin=c]{90}{\textsc{SimSiam}}} 
         & \textit{i.i.d.} & lr: 0.03 & lr: 0.01 &
          \multirow{5}{*}{\rotatebox[origin=c]{90}{\footnotesize\textsc{$b=10$, $n_p=3$}}}\\
         & \textit{CaSSLe} & lr: 0.03 & lr: 0.03 &\\
         & finetuning & lr: 0.01 & lr: 0.01  &\\
         & LUMP & lr: 0.03 & lr:  0.01  &\\
         & CLA-b & lr: 0.01, $\omega$: 0.1 & lr: 0.01, $\omega$: 1.0 &\\
         \cmidrule{2-4}
         & \textit{i.i.d.} & lr: 0.1 & lr: 0.03 &
           \multirow{7}{*}{\rotatebox[origin=c]{90}{\footnotesize\textsc{$b=30$, $n_p=1$}}}\\
         & \textit{CaSSLe-R} & lr: 0.1 & lr: 0.03 &\\
         & ER w/ reservoir & lr: 0.03 & lr: 0.03 &\\
         & ER w/ FIFO & lr: 0.03 & lr: 0.03 &\\
         & MinRed & lr: 0.03 & lr: 0.03 &\\
         & CLA-R & lr: 0.2, $\omega$: 0.3 & lr: 0.03, $\omega$: 1.0 &\\
         & CLA-E & lr: 0.2, $\omega$: 0.3 & lr: 0.1, $\omega$: 1.0 &\\
         \midrule
         \multirow{13}{*}{\rotatebox[origin=c]{90}{\textsc{SimCLR}}} 
        & \textit{i.i.d.} & lr: 0.01 & lr: 0.003 &
         \multirow{5}{*}{\rotatebox[origin=c]{90}{\footnotesize\textsc{$b=10$, $n_p=3$}}}\\
         & \textit{CaSSLe} & lr: 0.01 & lr: 0.01 &\\
         & finetuning & lr: 0.01 & lr: 0.003  &\\
         & LUMP & lr: 0.01 & lr: 0.003  &\\
         & CLA-b & lr: 0.01, $\omega$: 1.0 & lr: 0.01, $\omega$: 0.3 &\\  
         \cmidrule{2-4}
         & \textit{i.i.d.} & lr: 0.1 & lr: 0.03 &
         \multirow{7}{*}{\rotatebox[origin=c]{90}{\footnotesize\textsc{$b=30$, $n_p=1$}}}\\
         & \textit{CaSSLe-R} & lr: 0.03 & lr: 0.03 &\\
         & ER w/ reservoir & lr: 0.01 & lr: 0.01 &\\
         & ER w/ FIFO & lr: 0.01 & lr: 0.01 &\\
         & MinRed & lr: 0.01 & lr: 0.01 &\\
         & CLA-R & lr: 0.03, $\omega$: 0.3 & lr: 0.05, $\omega$: 1.0 &\\
         & CLA-E & lr: 0.03, $\omega$: 0.1 & lr: 0.05, $\omega$: 1.0 &\\
         \midrule
         \multicolumn{2}{c}{SCALE} & lr: 0.003 & lr: 0.003 \\
         \multicolumn{2}{c}{Osiris-R} & lr: 0.003 & lr: 0.003 \\
         \bottomrule
    \end{tabular}
\end{table*}
Table~\ref{tab:hyperparams_high} and Table~\ref{tab:hyperparams_low} detail the hyperparameters used in our experiments for both \textit{High} and \textit{Low CBP} settings, covering SimSiam- and SimCLR-based methods. We performed a grid search on a 10\% validation set to identify the optimal learning rate and, for the proposed CLA methods, also for the best regularization strength parameter, $\omega$, 
Compared to methods that utilize only replay samples without regularization (i.e. ER, MinRed), we observed that CLA can support larger learning rates, which can be attributed to the regularization effect introduced by the alignment loss. This trend of accommodating higher learning rates is also evident in other alignment-based methods, such as CaSSLe and CaSSLe-R.

\subsection{$\tau$ Analysis}
\begin{figure}[h]
    \centering
    \includegraphics[width=0.5\linewidth]{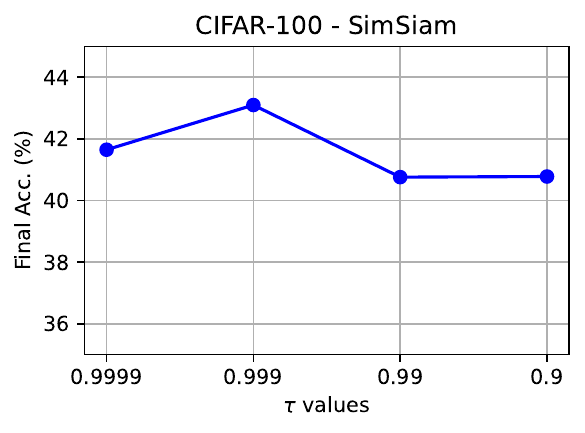}
    \caption{\textsc{Final Accuracy} of CLA-E on \textit{High CBP} CIFAR-100 with varying $\tau$ EMA hyperparameter, which controls $\theta'$ update speed, the EMA network used as target for the alignment. }
    \label{fig:tau}
\end{figure}

In Figure \ref{fig:tau} we report an ablation on the value of $\tau$, which is the hyperparameter that controls the update of the EMA network $\theta'$, as explained in Section \ref{sec:cla}.
Consistent with other results in representation learning literature \citep{grill2020byol, he2020moco}, we observed that the impact of $\tau$ is limited on the overall performance of the method.

\subsection{$\omega$ Analysis}
\begin{figure}[h]
    \centering
    \includegraphics[width=1.0\linewidth]{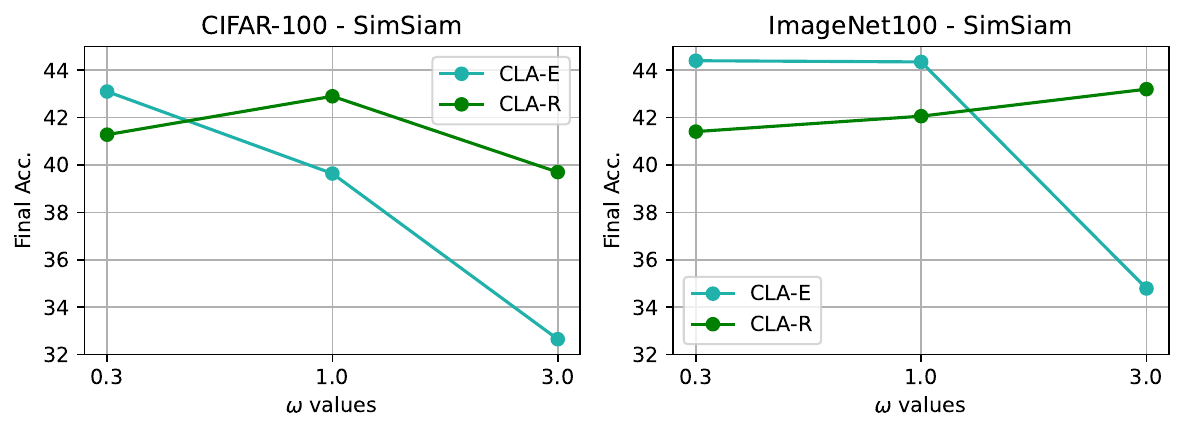}
    \caption{\textsc{Final Accuracy} of CLA-E and CLA-R on \textit{High CBP} CIFAR-100 with varying $\omega$, the alignment strength.}
    \label{fig:omega}
\end{figure}

Figure \ref{fig:omega} shows the variation of \textsc{Final Accuracy} in both CIFAR-100 and ImageNet100 with different values of $\omega$, the hyperparameter that controls CLA alignment strength.
While CLA-R is less sensitive to the choice of $\omega$, we can instead observe that CLA-E favors lower values of $\omega$, especially in CIFAR-100; higher values, such as $\omega=3.0$ cause a steep decrease in accuracy.

\end{document}